\RequirePackage{pgf}
\documentclass[pdflatex,sn-mathphys-num]{sn-jnl}
\usepackage{nicematrix}

\usepackage{amsmath, amssymb, amsfonts}  % Core AMS math
\usepackage{amsthm}                      % Theorem environments
\usepackage{mathrsfs}                    % Script-style math fonts

\usepackage{graphicx}                    % Image support
\usepackage{xcolor}                      % Color support

\usepackage{longtable}
\usepackage{array}                       % Enhanced tabular support
\usepackage{booktabs}                    % Better table lines
\usepackage{multirow}                    % Multirow cells (must come after array)
\usepackage{colortbl}                    % Colored table rows/columns

\usepackage{algorithm}                   % Algorithm float
\usepackage{algpseudocode}               % Pseudocode environment (uses algorithmicx internally)
\usepackage{listings}                    % Code listings (use instead of minted)

\usepackage{manyfoot}                    % Multiple footnote series (careful with footmisc)

\usepackage{textcomp}                    % Extra text symbols

\usepackage{float}                       % Enables [H] placement
\restylefloat{table}                     % Force table float styles

\usepackage[title]{appendix}             % Appendix formatting
\usepackage{makecell}
\usepackage{nicematrix}
\newcommand{\customsize}{\fontsize{8}{11}\selectfont}

\theoremstyle{thmstyleone}%
\theoremstyle{thmstyletwo}%

\theoremstyle{thmstylethree}%

\begin{document}

\title[Article Title]{LLM-Assisted Question-Answering on Technical Documents Using Structured Data-Aware Retrieval Augmented Generation}

\author{Shadman Sobhan\textsuperscript{1}, Dr. Mohammad Ariful Haque\textsuperscript{2*}}

\email{shadmansobhan114@gmail.com, arifulhoque@eee.buet.ac.bd}

% \author*[2]{Dr. Mohammad Ariful Haque}\email{arifulhoque@eee.buet.ac.bd}

\affil*[1,2]{Department of Electrical \& Electronic Engineering, Bangladesh University of Engineering \& Technology, Dhaka, 1200, Bangladesh}

% \title[Article Title]{LLM-Assisted Question-Answering on Technical Documents Using Structured Data-Aware Retrieval Augmented Generation}

% \author{Shadman Sobhan\textsuperscript{1}}
% \ead{shadmansobhan114@gmail.com}
% \author*{Dr. Mohammad Ariful Haque\textsuperscript{2}}
% \ead{arifulhoque@eee.buet.ac.bd}

% \address{\textsuperscript{1,2} Department of Electrical and Electronic Engineering,\\Bangladesh University of Engineering and Technology, Dhaka-1000, Bangladesh}

\abstract{Large Language Models (LLMs) are capable of natural language understanding and generation. But they face challenges such as hallucination and outdated knowledge. Fine-tuning is one possible solution, but it is resource-intensive and must be repeated with every data update. Retrieval-Augmented Generation (RAG) offers an efficient solution by allowing LLMs to access external knowledge sources. However, traditional RAG pipelines struggle with retrieving information from complex technical documents with structured data such as tables and images.
In this work, we propose a RAG pipeline, capable of handling tables and images in documents, for technical documents that support both scanned and searchable formats. Its retrieval process combines vector similarity search with a fine-tuned reranker based on Gemma-2-9b-it. The reranker is trained using RAFT (Retrieval-Augmented Fine-Tuning) on a custom dataset designed to improve context identification for question answering.
Our evaluation demonstrates that the proposed pipeline achieves a high faithfulness score of 94\% (RAGas) and 96\% (DeepEval), and an answer relevancy score of 87\% (RAGas) and 93\% (DeepEval).  Comparative analysis demonstrates that the proposed architecture is superior to general RAG pipelines in terms of table-based questions and handling questions outside context.

\textbf{Keywords:} LLM, Hallucination, Fine-tuning, RAG, Structured data, RAFT.}

\maketitle

\section{Introduction}\label{sec1}
Large Language Models (LLMs) demonstrate impressive capabilities in various natural language processing tasks. The introduction of the Transformer architecture by Vaswani et al. \cite{vaswani2017attention} marked a major breakthrough in Natural Language Processing (NLP). With  BERT, the bidirectional approach impacted the field of NLP, through which text could be read from both sides. GPT-3 by OpenAI demonstrated that LLMs can be applied in real-life tasks. Through GPT-3, LLMs began to be seen as general-purpose models capable of performing a variety of tasks.

However, LLMs face several limitations. Outdated training data can lead to wrong answers \cite{mousavi2024your}. One of the major problems of LLMs is "Hallucination" \cite{reddy2024hallucinations,banerjee2024llms}, which means inaccurate answers by LLMs that might seem true and contextually relevant \cite{augenstein2024factuality}. As general-purpose models, LLMs often fail to provide precise answers for domain-specific questions, such as those found in technical manuals or scientific reports. Also, LLMs cannot be asked questions about private data, as they lack access to the data.

There are various ways to solve the problems mentioned. The problem of outdated knowledge and hallucination can be solved by training LLMs regularly. LLMs can be made domain-specific by fine-tuning using a specialized dataset \cite{zhuang2023toolqa, huang2023dsqa}. But this is not a feasible solution always. As textual data is continuously increasing, fine-tuning only once is not enough. 

A promising solution to address these challenges of \textbf{hallucination}, \textbf{not being domain specific}, \textbf{private data}, \textbf{outdated training data} is \textbf{Retrieval Augmented Generation (RAG)}. In RAG, an LLM is given context through any form of textual data like Portable Document Format(PDF), Text File(TXT), Comma-Separated Values(CSV), Document(DOC), etc. The LLM generates answers based on the given context. 

However, LLMs are mostly trained on text-based data for information, not on data that can help them identify true and false contexts to answer a question from given contexts. So, even in RAG, there is a possibility that LLMs might not be able to identify the true context. 
To address the context identification issue in RAG, we utilized Retrieval-Augmented Fine-Tuning (RAFT), which explicitly teaches the model to distinguish between relevant and irrelevant contexts.

To summarize, the existing gaps in RAG pipelines that motivated us for this work are as follows:
\begin{itemize}
  \item RAG pipelines are mostly proposed for searchable PDFs, not for scanned documents.
  \item Generally, retrieval is done through vector similarity search, which might bring an irrelevant document.
  \item Existing pipelines do not work well when the document contains structured data like tables, images.
  \item Technical documents often include a lot of structured data. They can also be complex and hard to understand, making it difficult to find the exact answer to a question. Because of this, RAG systems often struggle to provide accurate answers from technical documents.
\end{itemize}

Our main contributions are as follows.
\begin{itemize}
    \item The proposed pipeline supports both scanned and searchable documents as inputs.
    \item For enhanced retrieval, we incorporated an LLM-based reranker along with traditional vector similarity search.
    \item For the LLM to work as a reranker, we fine-tuned it using a special method called "RAFT". 
    \item We created a custom dataset to apply the RAFT technique.
    \item The proposed method extracts and interprets structured data (e.g., tables, images) for improved question answering.
    \item The overall pipeline proposed in this article works exceptionally well for technical documents.
\end{itemize}

\begin{figure}[h!]
  \centering
  \begin{minipage}{1.2\textwidth}
    \centering
    \includegraphics[width=\linewidth]{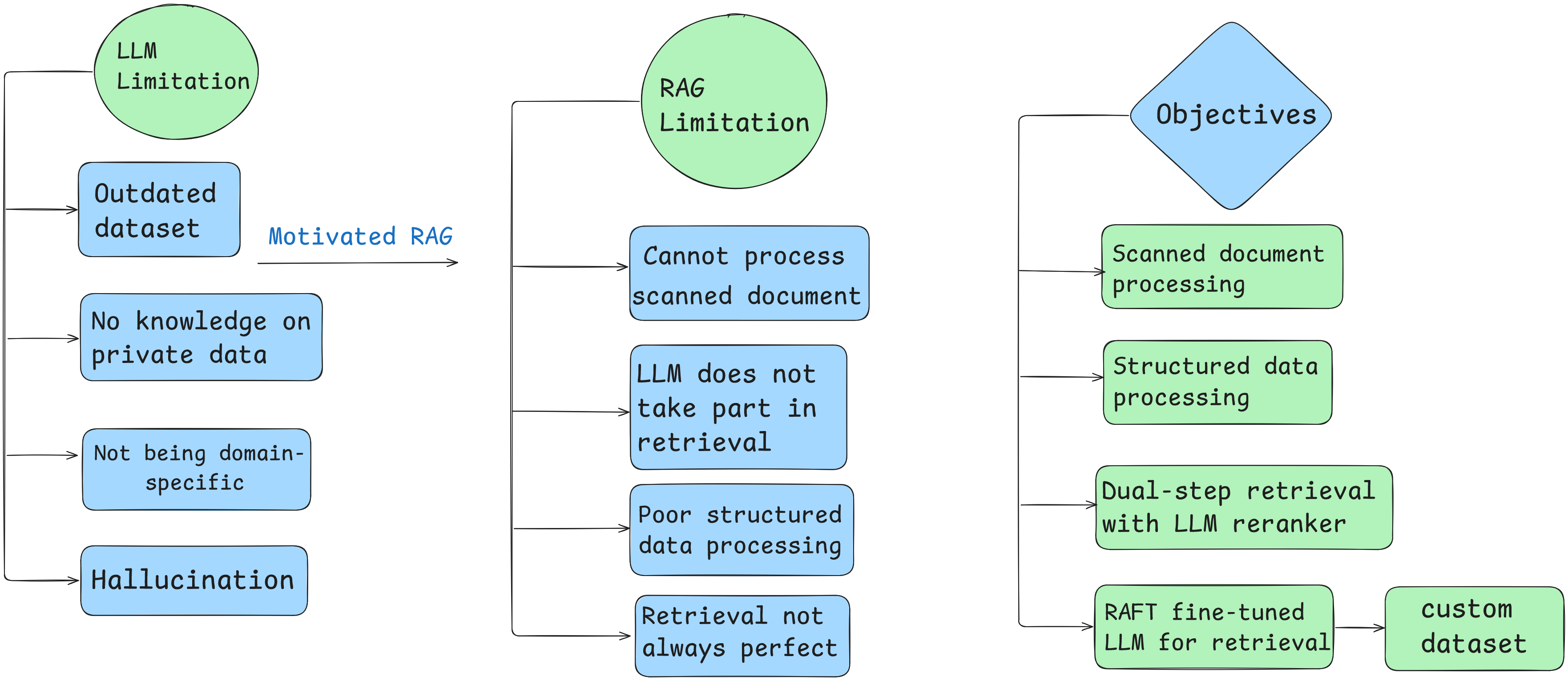}
    \caption{Motivations and Objectives}
    \label{fig:ga}
  \end{minipage}
\end{figure}

Figure \ref{fig:ga} clearly shows the motivations and objectives of this work.

The rest of the paper is structured as follows: Section \ref{sec2} \textbf{Literature Review} discusses the works that paved the way for this research. Section \ref{sec3} \textbf{Methodology} explains the techniques and procedures we employed in this work. After that section \ref{sec4} talks about the experimental setup and training methods used. Then section \ref{sec5} focuses on the evaluation procedure and section \ref{sec6} concludes this paper.

\section{Literature Review}\label{sec2}

The development of large language models (LLMs) has significantly advanced the field of Natural Language Processing (NLP) \cite{mohammad2023large}. Models such as BERT and GPT-1 laid the foundation for the development of many subsequent LLMs \cite{gaikwad2022extensive}.
BERT can read in both directions, unlike traditional models, which can read from either left to right or vice versa \cite{devlin2019bert}. This bidirectional mechanism allowed BERT to capture deeper contextual relationships  \cite{artetxe2022role}. Previous models had limited contextual awareness, but BERT changed that by making the relationships between words clearer \cite{ozccift2021advancing,wu2022study}.
Instead of training on labeled data (supervised learning), GPT-1, developed by Radford et al. \cite{radford2018improving}, was trained using unsupervised learning by predicting the next word in large corpora. Later models like GPT-2 and GPT-3 became much larger and more powerful \cite{radford2019language,brown2020language}. With billions of parameters and trained on vast datasets, they could generate human-like text with impressive fluency \cite{gao2023examining}. 

The capabilities of these models have enabled a wide range of applications. For example, BERT and its variations have become standards in tasks like text classification and sentiment analysis \cite{chen2024systematic,chkirbene2024large}. Their contextual understanding has made them very useful \cite{yunianto2020domain}.

GPT-3 expanded the capabilities of LLMs. Beyond standard NLP tasks, it can contribute to creative writings, generate code, and solve complex problems \cite{zhang2021commentary}. Its ability to generate realistic and context-aware language has made it helpful in content creation, including article writing, poetry, or brainstorming ideas \cite{matsui2024human}.

%%%%%%%%%%%%%%%%%%%%%%%%%%%%%%%%%%%%%%%%%%%%%%%
One of the most popular applications of LLMs is question answering. Early QA systems were often rule-based and struggled with context. GPT-2 improved that by generating more natural and accurate responses \cite{zoph2016neural,dhingra2016gated,olabiyi2019dlgnet}. Its ability to keep track of previous words helped maintain context and made its answers more thoughtful and coherent.

However, even the best language models sometimes get facts wrong, especially when asked about things outside their training data. That’s where Retrieval-Augmented Generation (RAG) comes in \cite{lewis2020retrieval}. It retrieves relevant documents or information and then uses that data to guide the model’s response. This makes answers more factually accurate and contextually relevant \cite{yao2023llm,ji2024llm,martino2023knowledge,li2024enhancing}.
%%%%%%%%%%%%%%%%%%%%%%%%%%%%%%%%%%%%%%%%%%%%%%%

Recent advancements in Retrieval-Augmented Generation (RAG) have enhanced the ability of large language models (LLMs) to generate factually accurate and context-aware responses by utilizing external knowledge sources. One of the foundational works, Atlas by Izacard et al. \cite{izacard2023atlas}, introduced a retrieval-aware encoder-decoder architecture that enables few-shot learning with retrieved evidence, setting a strong benchmark in open-domain QA. RePlug, proposed by Shi et al. \cite{shi2023replug}, explored decoupling the retrieval mechanism from the LLM during training, demonstrating that post-hoc retrieval adaptation can match or surpass end-to-end retrievers, offering flexibility in RAG pipelines. Siriwardhana et al. \cite{siriwardhana2023improving} proposed RAG-end2end that can adapt to a domain-specific knowledge base by updating all components of the external knowledge base during training.

Even though understanding texts has become possible due to LLMs, parsing information from table data, and images remains a problem in Natural Language based tasks. Models like TAPAS (Tabular Parser) \cite{herzig2020tapas} and LayoutLM (Layout-aware Language Model) \cite{xu2020layoutlm} are good at understanding table layouts and answering questions from tabular data. However, in a RAG system, tables often need to be extracted separately from the input document, making complex document understanding a challenge.

For flowcharts, combining computer vision with NLP helps recognize the steps and logic \cite{song2022graph}. Image-based information extraction often involves OCR and object detection to handle charts and annotated visuals.

Lastly, fine-tuning helps adapt large models to specific tasks. Techniques like supervised fine-tuning and transfer learning make models better at tasks like sentiment analysis or classification \cite{howard2018universal}. More efficient techniques like LoRA and PEFT reduce the computational load while maintaining performance \cite{hu2022lora,han2024parameter,houlsby2019parameter}. Retrieval-Augmented Fine-Tuning (RAFT) takes things further by using external data during training to teach LLM to identify the correct context \cite{zhang2024raft}. 

\section{Methodology}\label{sec3}
\begin{figure}[h!]
    \centering
    \includegraphics[width=\textwidth]{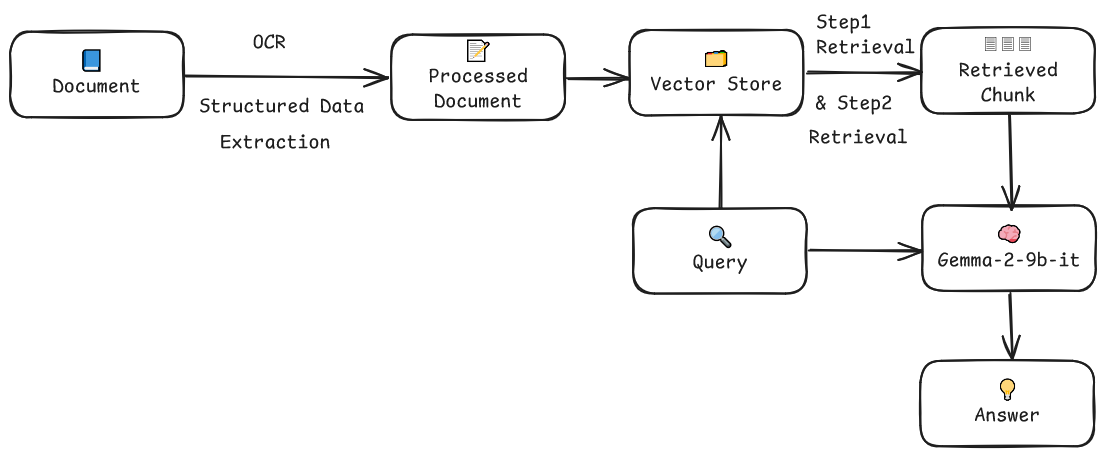}
    \caption{Methodology}
    \label{fig:MET}
\end{figure}

Figure \ref{fig:MET} illustrates the overall methodology of this work. The whole methodology can be divided into 3 parts: Input processing, Retrieval, and Answer generation.
% \begin{figure}[h!]
%     \centering
%     \includegraphics[width=1.2\textwidth]{Metho_ex(1).png}
%     \caption{Methodology}
%     \label{fig: MET}
% \end{figure}

\subsection{Document Processing:}
The pipeline proposed in this work can process both searchable and non-searchable(scanned) documents as input. General RAG pipelines and processing in NLP are suitable for searchable documents, as they can be converted into vector representations. Therefore, we shall convert scanned documents to searchable documents first. Besides, embeddings cannot represent structured data properly. So we will also extract tables and images from the given document, then express them in a format suitable for further processing.

\begin{figure}[h!]
    \centering
    \includegraphics[width=\textwidth]{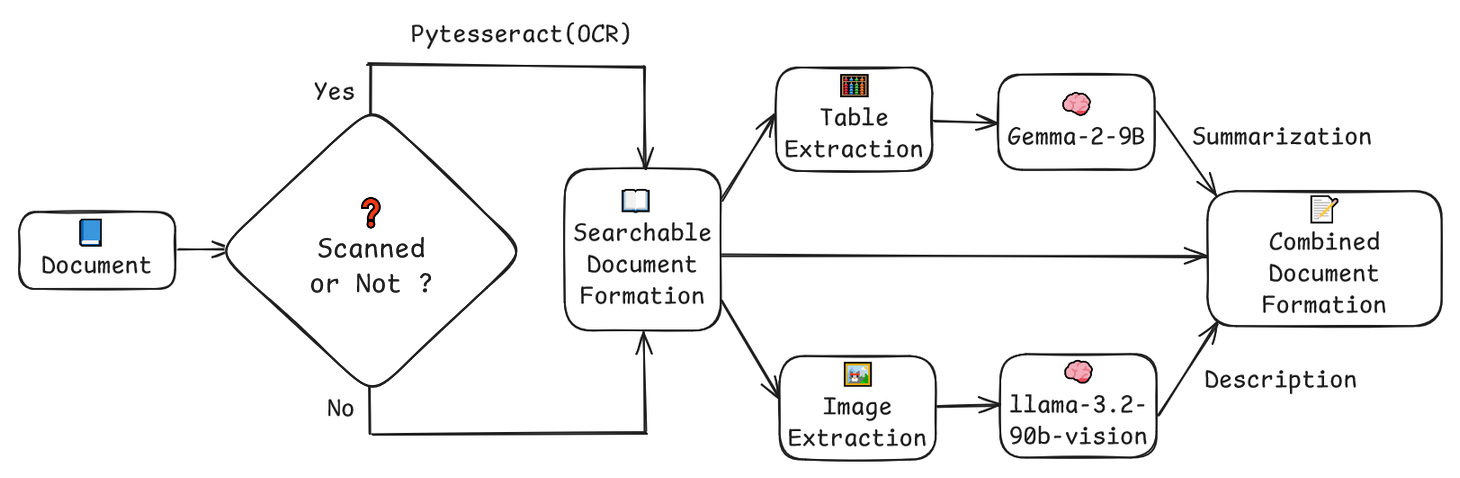}
    \caption{Extraction of structured data}
    \label{fig:Document Processing}
\end{figure}

Figure \ref{fig:Document Processing} shows the process of forming a searchable document and separating tables and images from it. 

\subsubsection{Searchable Document Formation:}

Our pipeline converted scanned documents into a searchable format using Optical Character Recognition (OCR). This ensured that the text within the document could be indexed and queried effectively.

The first step was to determine whether the document was a scanned document. Scanned documents typically contain images of text rather than selectable text data, making them non-searchable. When a document was identified as a scanned document, the Python library \textbf{Pytesseract} was used to apply OCR and convert the images of text into a searchable PDF format.

Algorithm \ref{alg:OCR} shows the procedure.

\begin{algorithm}[H]
\caption{OCR-Based Scanned Document Conversion}
\label{alg:OCR}
\begin{algorithmic}[1]
\For{each image of a PDF page}
    \State Perform OCR on the image, producing a PDF with recognized text.
    \State Save the OCR result as a temporary PDF file.
    \State Open the temporary PDF and insert it into the new document.
    \State Close and delete the temporary PDF file.
\EndFor
\end{algorithmic}
\end{algorithm}

\subsubsection{Why is Table Parsing Needed?}

In RAG, the texts within a document are converted to embeddings. In that case, borders that are critical to table structure are not held within the vector store, and so it becomes difficult to understand tables. For example, if Table \ref{tab:Tab1} is present in a document, the table will be seen by LLMs (through vector representation) as in Table \ref{tab:Tab2}.


% \usepackage{multirow}
\begin{table}[htbp]
\centering
\caption{Actual Table}
\label{tab:Tab1}
\begin{NiceTabular}{|l|l|l|}
\hline
\multirow{2}{*}{Name} & \multicolumn{2}{l|}{Activity} \\ 
\cline{2-3}
                      & Morning & Night \\ \hline
A & Class & Study \\ \hline
B & Office & Rest \\ \hline
\end{NiceTabular}
\end{table}

\begin{table}[H]
\centering
\caption{Table In Vector Database}
\label{tab:Tab2}
\begin{tabular}{lll}
\multirow{2}{*}{Name} & \multicolumn{2}{l}{Activity} \\
                      & Morning        & Night       \\
A                     & Class          & Study       \\
B                     & Office         & Rest       
\end{tabular}
\end{table}

In such cases, it becomes difficult to relate 'night' to 'study' or 'rest'. A further problem appears when one column holds two or more sub-columns, as in the mentioned table. In that case, an LLM cannot distinguish between column headers and column elements. 

This issue can be mitigated by converting the table into a descriptive sentence format, such as ''\textit{Name A activity class in the morning and study at night. Name B activity office in the morning and rest at night.}''

\subsubsection{Table Data Parsing:}
Each page of the input document was converted to images. By using a pre-trained YOLO model available in the Unstructured library, tables were identified from each page. Then the extracted tables were converted to HTML code through the Unstructured library, and then LLM was used to summarize that. The prompt for summarization is as follows: 

\textbf{\textit{"You will thoroughly explain the numbers/ row components of table \{context\} with every detail. It is important to mention every number. Do not avoid any row. Suppose the column heads are 'Person' and 'Hobby. ' And a row is 'A', 'Sleep'. So you will write it as Person A's hobby is to sleep."}} 

Here, the context refers to the extracted HTML code.

In this work, we used \textbf{"gemma2-9b-it"} from GROQ for the summarization of HTML code, as it strictly maintained the prompt compared to other LLMs from GROQ. The procedure to extract table information from the input document is given in Algorithm~\ref{alg:Table_Algo}.

\begin{algorithm}[h] % 'H' forces the algorithm to stay here
\caption{Table Data Extraction and Description Pipeline}
\label{alg:Table_Algo}
\begin{algorithmic}[1]

\Require Input document $D$
\Ensure Descriptions of extracted tables saved to a separate PDF

\State Convert all pages of $D$ into separate images $I_1, I_2, \dots, I_n$

\For{each image $I_i$}
\State Apply YOLO to detect and extract tables $T_{i1}, T_{i2}, \dots, T_{im}$
\EndFor

\State Convert each extracted table $T_{ij}$ into HTML code $H_{ij}$

\For{each HTML table $H_{ij}$}
\State Use language model $L$ to generate description $D_{ij}$
\EndFor

\State Save all descriptions $D_{ij}$ to a separate PDF document

\Return All descriptions $D_{ij}$

\end{algorithmic}
\end{algorithm}

\subsubsection{Image Data Parsing:}
Images were separated from the document using pre-trained YOLO from the Unstructured library. After separating images from a document, an effective approach to understand them was to use VLMs (Vision Language Models), which could describe visual information through text description, along with the capability of understanding texts.

The prompt for image description is: 

\textbf{\textit{"This image will be used in Retrieval Augmented Generation. So explain it as much detail as you can."}}

For explaining images, multimodal LLM \textbf{"llama-3.2-90b-vision-preview"} from GROQ was used.

\subsubsection{Combined Document Formation}
After extracting table and image information, three documents were kept in a single directory: a PDF containing the description of table data, a PDF containing the description of image data, and the base document. As the directory carried all the information present in the structured data, but in a form suitable for storing in a vector database, it could be used for ingestion.
In our system, the Simple Directory Reader was employed to load documents from that directory, enabling quick ingestion of text data into our pipeline for further analysis and embedding generation.

\subsection{Enhanced Retrieval:}

\begin{figure}[h!]
    \centering
    \includegraphics[width=1\textwidth]{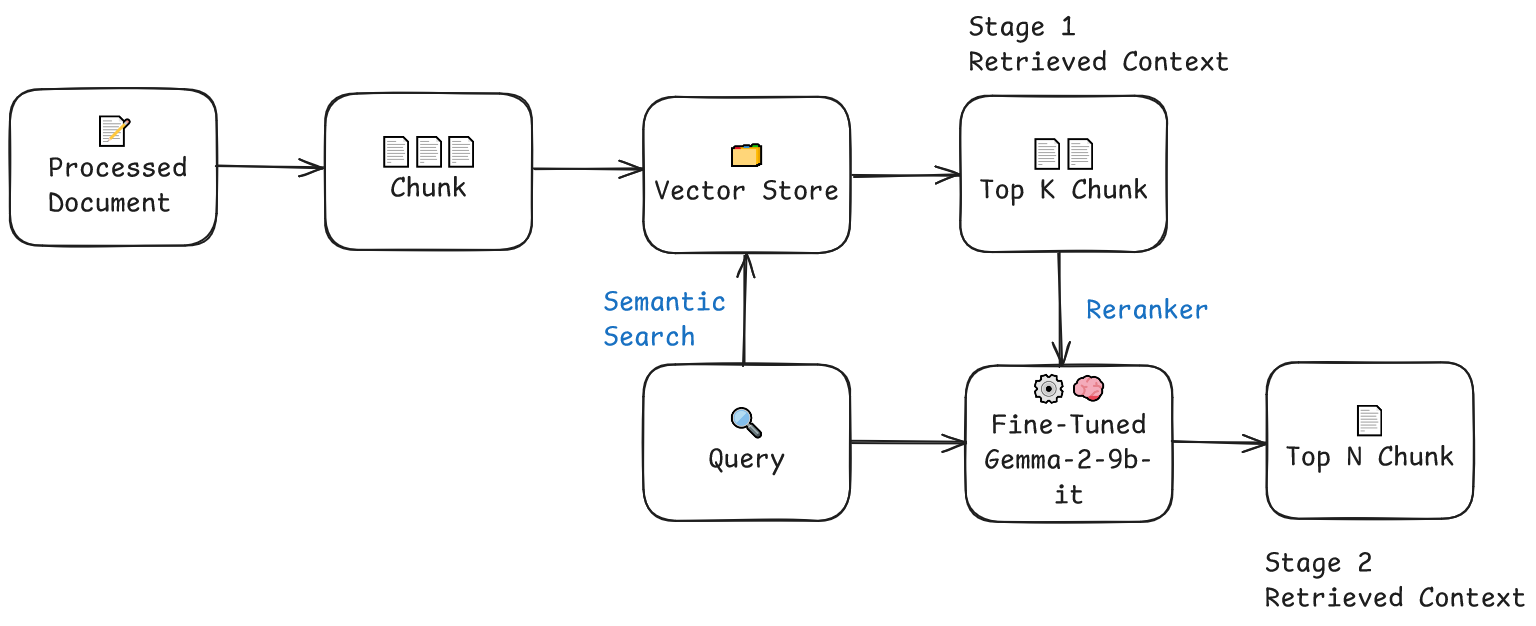}
    \caption{Retrieval of contexts}
    \label{fig:Retrieval of contexts}
\end{figure}

Figure \ref{fig:Retrieval of contexts} shows the process of retrieving relevant contexts from the given combined document.

\subsubsection{Stage 1 Retrieval- Semantic Search:}
At first, the document was chunked. Chunking is a technique where large amounts of information are broken down into smaller pieces, or "chunks," to make it easier for information retrieval. Once the information is divided into smaller, manageable chunks, each word of the chunk is transformed into a vector representation using embedding models. 

In this work, we chunked the input document using fixed-size chunks of 512 tokens. Then we used the \textbf{BAAI/bge-small-en-v1.5} model from Hugging Face as vector embeddings. The resulting vector representations were then stored in a Vector Store. We used \textbf{FAISS} in this work as vector store. When a user submitted a query, the system first processed the query by removing stop words — common but often unnecessary words like "and," "the," or "is". The cleaned query was then converted into its vector representation, which was compared to the stored vectors in the vector database to find the most similar contexts/chunks.

From this search, the top K most relevant contexts were retrieved — in our case, we set K=10. These contexts are then passed to the next stage of the retrieval process.

In most Retrieval Augmented Generation (RAG) systems, context retrieval relies solely on this method, known as \textbf{semantic search}. 

\subsubsection{Stage 2 Retrieval- LLM Reranker:}
The initial retrieval step brought back the top K most relevant contexts based on semantic similarity. However, these results might still contain noise or less relevant information. 

An LLM reranker is a powerful model designed to refine and reorder the retrieved contexts by evaluating their relevance to the user’s query. Unlike simple similarity measures, \textbf{the reranker uses the capabilities of the LLM to assess not just surface-level similarity but the true semantic alignment between the query and each context retrieved in the first step}. That is the reason reranker was used in this study, as it would bring contexts perfectly relevant to the question. 

Once the initial K results were retrieved, the reranker selected the top N contexts and ranked them in order of relevance, ensuring that the most useful and accurate information appears at the top. This additional step enhances the quality of context passed to the LLM for generating responses, leading to more precise and informative outputs. We selected N=3.

The reason for setting N=3 is that an exact answer might lie within two chunks, so we set a value of \(N>1\). Besides, LLM reranker ranks the retrieved contexts. So, even though table data is present in both the document described by LLM and the base document, as the information is clearer in the document described by LLM, it is chosen for answer generation.

\subsection{Text Generation:}

The generation pipeline in the Retrieval-Augmented Generation (RAG) framework is responsible for transforming the retrieved context into coherent, informative, and contextually aware responses. Once the retrieval component has fetched the most relevant documents or knowledge snippets, the generation pipeline uses this information to generate human-like answers.

In our implementation, we utilized \textbf{fine-tuned 'Gemma-2-9b-it-GPTQ'}. The generation pipeline took the query and the retrieved context as input and constructed a well-formed response. 

\begin{figure}[h]
    \centering
    \includegraphics[width=1\textwidth]{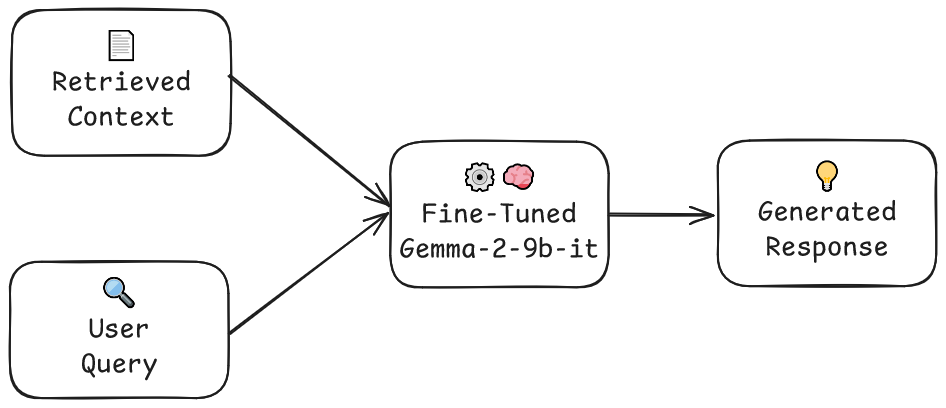}
    \caption{Text Generation}
    \label{fig: TG}
\end{figure}

Figure \ref{fig: TG} illustrates the text generation methodology.

\section{Experimental Setup \& Training}\label{sec4}

\subsection{Utilized Models:}
The models we utilized in this study, along with proper reasoning, are below in Table \ref{tab:Models}
\begin{table}[H]
\caption{Models used in this study}
\label{tab:Models}
\resizebox{\textwidth}{!}{%
\begin{tabular}{|l|l|l|l|}
\hline
Purpose                                                               & \begin{tabular}[c]{@{}l@{}}Which part this \\ was used\end{tabular} & Model / LLM                                                               & Reason                                                                                                                                                             \\ \hline
Embeddings                                                           & All parts                                                           & BAAI/bge-small-en-v1.5                                                    & \begin{tabular}[c]{@{}l@{}}Compact Size, faster \\ interference, open source.\end{tabular}                     \\ \hline
Vector Store                                                         & All parts                                                           & FAISS                                                                     & \begin{tabular}[c]{@{}l@{}}Optimized for large-scale \\ search, open source, efficient \\ indexing, and GPU support \\ for faster indexing.\end{tabular}           \\ \hline
OCR                                                                  & RAG pipeline                                                        & Pytesseract                                                               & \begin{tabular}[c]{@{}l@{}}High speed and \\ accuracy, open source\end{tabular}                                                                                    \\ \hline
\begin{tabular}[c]{@{}l@{}}LLM for Table \\ Explanation\end{tabular} & RAG pipeline                                                        & \begin{tabular}[c]{@{}l@{}}Gemma-2-9b-it\end{tabular}        & \begin{tabular}[c]{@{}l@{}}Strictly follows Prompt \\ compared to other \\ free LLMs\end{tabular}                                                                  \\ \hline
\begin{tabular}[c]{@{}l@{}}VLM for Image \\ Description\end{tabular} & RAG pipeline                                                        & \begin{tabular}[c]{@{}l@{}}Llama-3.2-90b-vision\end{tabular} & \begin{tabular}[c]{@{}l@{}}Open source and \\ detail description provider\end{tabular}                               \\ \hline
Document Read                                                        & RAG pipeline                                                        & Simple Directory Reader                                                   & \begin{tabular}[c]{@{}l@{}}Simplest processing, no \\ need to merge documents\end{tabular}                                                                         \\ \hline
LLM                                                                  & RAG pipeline                                                        & \begin{tabular}[c]{@{}l@{}}Gemma-2-9b-it\\ Fine Tuned\end{tabular}        & \begin{tabular}[c]{@{}l@{}}Precise answering, no \\ unnecessary explanation. \\ Different users prefer \\ different LLMs.\end{tabular}                             \\ \hline
Reranker                                                             & RAG pipeline                                                        & \begin{tabular}[c]{@{}l@{}}Gemma-2-9b-it\\ Fine Tuned\end{tabular}         & \begin{tabular}[c]{@{}l@{}}Using a different LLM \\ would require more \\ resources, making \\ it not suitable for a \\ resource-constrained setting.\end{tabular} \\ \hline
\begin{tabular}[c]{@{}l@{}}RAFT Dataset \\ Preparation\end{tabular}  & RAFT                                                                & \begin{tabular}[c]{@{}l@{}}Qwen-2.5-32b\\ from GROQ\end{tabular}          & \begin{tabular}[c]{@{}l@{}}Very high context \\ window size compared to \\ other free LLMs. Detail\\ in \ref{DP}\end{tabular}                                                            \\ \hline
LLM Evaluation                                                       & Evaluation                                                          & \begin{tabular}[c]{@{}l@{}}Llama-3.3-70B\\ from GROQ\end{tabular}         & \begin{tabular}[c]{@{}l@{}}Different LLMs\\ produce different \\ results. It is up to\\ the evaluator.\end{tabular}                                                \\ \hline
\end{tabular}}
\end{table}

\subsection{RAFT- Retrieval-Augmented Fine-Tuning}
Retrieval-Augmented Generation (RAG) helps language models by retrieving external information before answering. However, models often struggle to find the correct context. RAFT (Retrieval-Augmented Fine-Tuning) combines both approaches. It trains models with both relevant and irrelevant contexts, helping them learn what to focus on. RAFT trains rerankers to identify the best context according to the user query.

\subsubsection{Components of RAFT:}
In RAFT, each data in the training dataset consists of:
\begin{itemize}
    \item A question (Q)
    \item A set of documents (Dk), which are broken down into two types:
    \begin{itemize}
        \item \textbf{“Oracle”} documents — documents that contain the answer to the question. 
        \item \textbf{“Distractor”} documents — documents that do not contain relevant information for answering the question.
    \end{itemize}
    \item A chain-of-thought style answer (A*): An answer generated from the oracle documents that includes a detailed reasoning process.
\end{itemize}

\subsubsection{Dataset Preparation:}
\label{DP}
We initially selected three publicly available technical manuals: the Circuit Breaker Testing Guide [\href{https://s3-eu-west-1.amazonaws.com/productdatasheetsv2/Circuit%20Breaker%20Testing%20Guide_3586.pdf}{link}], Power Cable Testing [\href{https://www.cedengineering.com/userfiles/Power%20Cable%20Testing-R1.pdf}{link}], and Power Transformer Testing [\href{https://www.cedengineering.com/userfiles/Power%20Transformer%20Testing-R1.pdf}{link}]. These manuals were combined into a single document, which is available on GitHub [\href{https://github.com/ShadmanSobhan/SDA-RAG/blob/main/Testing%20Manual%20for%20RAFT%20QA%20Generation.zip}{link}]. The combined file was split into 240 chunks for easier processing. Since this was not enough data for this study, we added more content, including other technical manuals of engines, textbooks, and fiction books such as the Harry Potter series. These additional texts were split into 800 chunks, bringing the total number of chunks to 1,040 (technical documents chunk 740, general document chunk 300). General documents were included to enhance the \textbf{generalizability} of the fine-tuned model.

From each chunk, exactly 1 question was generated by LLM. That chunk was the relevant chunk or oracle in that case. Randomly, 2 other chunks were selected that were labeled as false context. LLM generated a CoT(Chain of Thought) style answer from the oracle and question. By this method, 1 question, 1 relevant context or oracle, 2 distractor contexts to distinguish the correct context from irrelevant contexts, and 1 answer with CoT(Chain of Thought) was generated from each chunk. 

\begin{figure}[h!]
    \centering
    \includegraphics[width=1\textwidth]{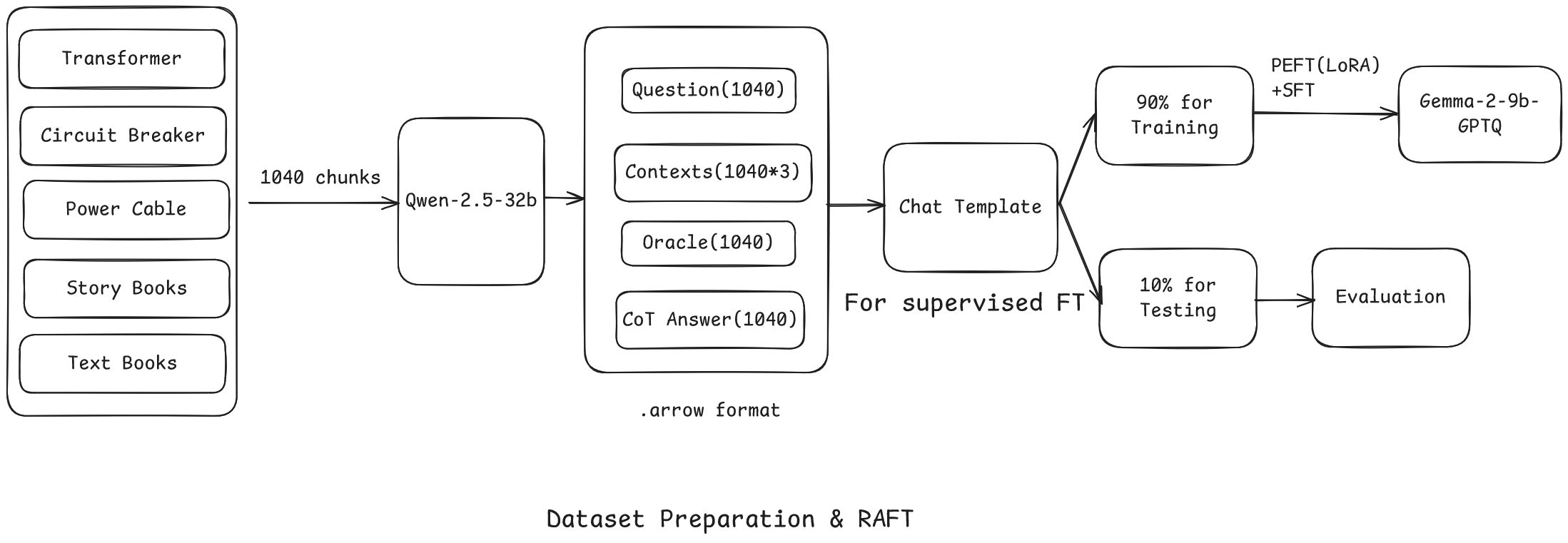}
    \caption{RAFT Dataset Preparation and Fine-Tuning Process}
    \label{fig:LLMF}
\end{figure}

For LLM, we used \textbf{"Qwen-2.5-32b"} from GROQ. The reason for using this specific LLM is its context window size of 128k. The 128k context window of Qwen-2.5-32b allowed us to generate longer, more contextually grounded CoT responses, outperforming shorter-context models like Mixtral-8x7b and Gemma-2-9b. For embeddings, we used \textbf{"BAAI/bge-small-en-v1.5"}. The preparation of the dataset and fine-tuning methodology is shown in Figure \ref{fig:LLMF}

\subsection{Fine-Tuning Technique:}

For fine-tuning, we applied PEFT(LORA) and Supervised Fine-Tuning. Instead of updating the entire set of model parameters, PEFT modifies a small subset or introduces lightweight modules, significantly reducing the number of parameters that need to be trained. Methods like LoRA (Low-Rank Adaptation), Prefix Tuning, and Adapter Layers are popular PEFT strategies. Instead of updating all the parameters of a model, LoRA introduces small, trainable, low-rank matrices into the architecture. During fine-tuning, only these added matrices are optimized, while the original model weights remain frozen. 

SFT (Supervised Fine-Tuning) is a training approach where a pre-trained model is adapted to a specific task using labeled data. In SFT, the model learns by mapping inputs to their corresponding outputs, guided by a clear set of correct answers.

In this work, for fine-tuning, we proceeded as algorithm \ref{alg:LoRA_SFT}.

\begin{algorithm}[H] % 'H' forces the algorithm to stay here
\caption{LoRA Fine-Tuning and SFT Training Pipeline}
\label{alg:LoRA_SFT}
\begin{algorithmic}[1]

\Require Pre-trained model $M$, Prompt-based dataset $D$
\Ensure Fine-tuned model $M_{FT}$

\State Load the pre-trained Gemma-2-9b-it model $M$

\State Prepare the prompt-based training dataset $D$

\State Split dataset $D$ into training set $D_{train}$ and testing set $D_{test}$

\State Apply LoRA

\State Initialize the SFT Trainer $SFT$

\State Train the model using $SFT$ on $D_{train}$, optimizing LoRA parameters

\State Save the fine-tuned LoRA parameters $P_{LoRA}$

\State Combine $P_{LoRA}$ with the base model $M$ to create the final fine-tuned model $M_{FT}$

\State Evaluate $M_{FT}$ on $D_{test}$ and measure performance metrics

\Return Fine-tuned model $M_{FT}$

\end{algorithmic}
\end{algorithm}

Our given prompt to convert the arrow format dataset suitable for supervised fine-tuning was the following:  

\textbf{
\texttt{
\%\% for message in messages \%\% \\
\%\% if message['role'] == 'user' \%\% \\
\{\{ '<|user|>\textbackslash n' + message['content'] + eos\_token \}\} \\
\%\% elif message['role'] == 'system' \%\% \\
\{\{ '<|system|>\textbackslash n' + message['content'] + eos\_token \}\} \\
\%\% elif message['role'] == 'assistant' \%\% \\
\{\{ '<|assistant|>\textbackslash n' + message['content'] + eos\_token \}\} \\
\%\% endif \%\% \\
\%\% if loop.last and add\_generation\_prompt \%\% \\
\{\{ '<|assistant|>' \}\} \\
\%\% endif \%\% \\
\%\% endfor \%\%
}
}

\subsection{Fine-Tuned Weights Loading:}
The process began by loading Gemma-2-9b and its associated tokenizer from the Hugging Face model hub. The model is then quantized by using a 4-bit precision to reduce the computational requirements and memory footprint. Then the model was loaded, and a text generation pipeline was set up using the fine-tuned weights and tokenizer. This pipeline was configured with parameters such as the maximum number of tokens to generate, sampling methods, and controls for randomness and repetition. The weights are available in \href{https://drive.google.com/file/d/1bHZyvIBoz1XpzPAczgBVv0ByILO6XmDY/view?usp=sharing}{Google Drive}.

\subsection{Machine Configuration:}
The experimental workflow for this study is divided into four key segments: RAFT dataset preparation, large language model (LLM) fine-tuning, Retrieval-Augmented Generation (RAG) implementation, and RAG evaluation. All of these steps were conducted on Kaggle. Kaggle provides access to powerful GPUs like the NVIDIA Tesla P100, which was used in this study. 

\subsection{Hyperparameter Tuning:}

There are two main hyperparameters in Dataset Preparation: \texttt{num\_questions\_per\_chunk} and \texttt{num\_distract\_docs}, which are set to 1 and 2, respectively. 
For LLM Fine-Tuning, the dataset was split into 90:10 for training and testing. For LoRA, our selection of hyperparameters is as follows:
\begin{itemize}
    \item \textbf{r=8}: This is the rank of the LoRA (Low-Rank Adaptation) matrices. 

    \item \textbf{lora\_alpha=32}: The scaling factor for the LoRA updates. 

    \item \textbf{lora\_dropout=0.05}: The dropout rate applied to the LoRA layers. 

\end{itemize}

For Training Arguments, our configuration is as follows:
\begin{itemize}
    \item \textbf{fp16=True}: Enables half-precision floating point (FP16) training, which reduces memory usage and speeds up computation. 

    \item \textbf{learning\_rate=2.0e-05}: Sets the learning rate to $2 \times 10^{-5}$.

    \item \textbf{num\_train\_epochs=3}: Specifies that training will run for 3 epochs.

\end{itemize}

For the \texttt{SFTTrainer}, the key hyperparameters are:
\begin{itemize}

    \item \textbf{max\_seq\_length=2048}: Maximum sequence length for the input data.
\end{itemize}

For LLM Quantization Configuration:

\begin{itemize}
    \item \textbf{bits=4}: Specifies the number of bits used for quantization. 

\end{itemize}

For Text Generation Pipeline:

\begin{itemize}

    \item \textbf{max\_new\_tokens=500}: Sets the maximum number of tokens the model can generate in a single output.

    \item \textbf{temperature=0.01}: Controls the randomness of predictions; lower values make outputs more deterministic.
    \item \textbf{top\_p=1}: Implements nucleus sampling, where the model considers the top \(p\) cumulative probability mass for token selection.
    \item \textbf{top\_k=5}: Restricts token sampling to the top 5 most likely next tokens, enhancing coherent text generation.
    \item \textbf{repetition\_penalty=1.1}: Penalizes repeated tokens to reduce redundancy in the generated output.
\end{itemize}

\section{Results}\label{sec5}
\subsection{Structured Data Extraction}
The first part of our RAG pipeline is structured data extraction. This part includes both table and image extraction and explanation. 
\subsubsection{Table Data Extraction}

\begin{table}[h!]
\centering
\caption{Extracted Table Using Proposed Algorithm from a Document}
\label{tab:Sec-Table}
\begin{tabular}{|c|c|c|}
\hline
\textbf{Type of Insulation} & \textbf{$k$} & \textbf{$\tan \delta$} \\
\hline
Impregnated paper & 3.5 & $2.3 \times 10^{-3}$ \\ \hline
Impregnated PPP & 2.7 & $0.7 \times 10^{-3}$ \\ \hline
PVC & 5.8 & $8.0 \times 10^{-3}$ \\ \hline
XLPE & 2.3 & $0.1 \times 10^{-3}$ \\ \hline
HDPE & 2.3 & $0.1 \times 10^{-3}$ \\ \hline
EPR & 2.8 & $3.5 \times 10^{-3}$ \\
\hline
\end{tabular}
\end{table}

\textbf{The parsed information is: }

\textit{Row 1:}

\textit{Type of Insulation: Impregnated paper}

\textit{k : 3.5  This represents the thermal conductivity (k) of impregnated paper. Thermal conductivity is a measure of how easily heat can pass through a material. A higher k value means the material conducts heat more readily.}

\textit{$\tan\delta$: $2.3 \times 10^{-3}$ This likely refers to the tangent of the dissipation angle 
($\tan\delta$), a measure of dielectric loss in the material.}  

\textit{Row 2:}

\textit{Type of Insulation: Impregnated PPP}

\textit{k :2.7  This is the thermal conductivity of impregnated PPP (likely Polypropylene).}

\textit{$\tan\delta$: $0.7 \times 10^{-3}$ This is the tangent of the dissipation angle for impregnated PPP.}

\textit{Row 3:}

\textit{Type of Insulation: PVC
k : 5.8  This is the thermal conductivity of Polyvinyl Chloride (PVC).
$\tan\delta$: $8 \times 10^{-3}$ This is the tangent of the dissipation angle for PVC.
}
\textit{Row 4:}

\textit{Type of Insulation: XLPE
k : 2.3  This is the thermal conductivity of Cross-linked Polyethylene (XLPE).
$\tan\delta$: $0.1 \times 10^{-3}$ This is the tangent of the dissipation angle for XLPE.
}
\textit{Row 5:}

\textit{Type of Insulation: HDPE
k : 2.3  This is the thermal conductivity of High-Density Polyethylene (HDPE).
$\tan\delta$: $0.1 \times 10^{-3}$ This is the tangent of the dissipation angle for HDPE.}

\textit{Row 6:}

\textit{Type of Insulation: EPR
k : 2.8  This is the thermal conductivity of Ethylene Propylene Rubber (EPR).
$\tan\delta$: $3.5 \times 10^{-3}$ This is the tangent of the dissipation angle for EPR.
}

\subsubsection{Image to Text}
\begin{figure}[h!]
    \centering
    \includegraphics[width=0.6\textwidth, height=0.5\textwidth]{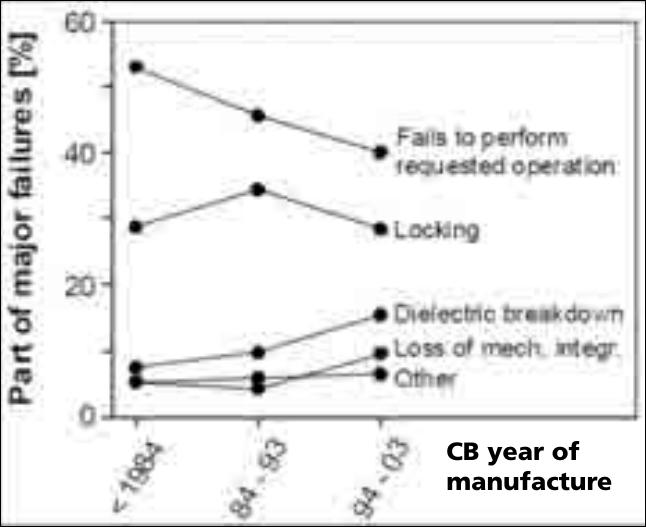}
    \caption{Image from a document}
    \label{fig:Extraction of image}
\end{figure}

Figure \ref{fig:Extraction of image} shows an image present in a certain document. The image was extracted using the proposed pipeline, and the information described by the VLM is given after the image.

The retrieved information is :

\textit{The image presents a line graph illustrating the percentage of major failures in CB manufacturing over time, categorized by type. The x-axis represents the calendar year of manufacture, spanning from 1984 to 2003, while the y-axis displays the percentage of major failures, ranging from 0\% to 60\%.}

\textit{Failure Categories:}

\textit{Fails to perform requested operation}

\textit{Locking}

\textit{Dielectric breakdown}

\textit{Loss of mech. Integr.}

\textit{Other}

\textit{Each category is represented by a distinct line on the graph, allowing for a clear comparison of the relative prevalence of each type of failure over time.}

\textit{Key Observations:}

\textit{Fails to perform requested operation}\textit{: This category consistently exhibits the highest percentage of major failures throughout the period, with no discernible trend or pattern.}

\textit{Locking}: \textit{Locking issues appear to be relatively stable, with a slight increase in the mid-1990s.}

\textit{Dielectric breakdown}: \textit{Dielectric breakdowns show a steady decline over the years, with a notable drop in the late 1990s.}

\textit{Loss of mech. Integr.}\textit{: This category remains relatively constant, with minor fluctuations.}

\textit{Other}: \textit{The "Other" category displays a slight downward trend, although it remains relatively stable overall.
}

\textit{Insights:}

\textit{The graph suggests that} \textit{fails to perform requested operation}\textit{has been a persistent issue in CB manufacturing, with no significant improvements over time.}

\textit{Dielectric breakdowns} \textit{have decreased substantially since the late 1990s, indicating potential advancements in materials or design.}

\textit{Locking issues}\textit{have been relatively stable but may warrant further investigation to prevent potential problems.}

\textit{Loss of mechanical integrity}\textit{and }\textit{other}\textit{categories have remained relatively consistent, with minor fluctuations.}

\textit{In conclusion, the graph provides valuable insights into the trends and patterns of major failures in CB manufacturing over the past two decades. By analyzing these trends, manufacturers can identify areas for improvement and implement targeted strategies to enhance product reliability and performance.
}

\subsubsection{Discussion of Structured Data Extraction}
The summarized texts explain the separated tables and images. From our evaluation, the proposed pipeline can successfully extract all the tables and images from a document. The summarized text of a table is always to the point and precise.

On the other hand, we used VLMs to explain images, specifically Llama 3.2-90B vision. As VLMs still do not perform best, their extracted info has a few problems, like sometimes they might describe the images' components with unnecessary descriptions rather than specific details.

Extracting structured data helps our RAG pipeline to answer more correctly, which will be evident from the following subsections.

\subsection{LLM Fine-Tuning}
The training performance while fine-tuning is given in Table \ref{tab:ft}

\begin{table}[h]
\centering
\caption{Fine-Tuning Log}
\label{tab:ft}
\begin{tabular}{|l|l|l|}
\hline
Epoch & Training Loss & Validation Loss \\ \hline
1     & No log        & 2.4845          \\ \hline
2     & 2.554         & 2.43544         \\ \hline
3     & 2.4799        & 2.425246        \\ \hline
\end{tabular}
\end{table}

\subsection{RAG Pipeline Evaluation}
Evaluating a Retrieval-Augmented Generation (RAG) system can be difficult because the quality of the answers depends on personal judgment. Different people may think differently about what makes a good answer. Still, we used some methods to measure how good the answers from our system are. These methods use large language models (LLMs) to score the answers, but this is also subjective, as different LLMs may give different scores. Also, some questions might get very high scores, while others may get low scores, depending on the type of given document and, clarity of the query. To make things more complicated, different research papers use different sets of questions and answers, so it’s hard to compare them directly.

In our work, we created a \textbf{test set of 50 questions} taken from different types of documents(35 from technical documents and 15 from general documents). This helped us test the system in a wide range of situations.

Evaluation was conducted using two tools: \textbf{RAGAS and DeepEval}, to check both how well the system retrieved documents and how well it generated answers. These tools needed an LLM to do the evaluation, and we used \textbf{LLaMA 3.3-70B} for this purpose. 

\subsubsection{Test Set Formation:}
 The test set contains questions, ground truth answers (taken exactly from the source documents), and the answers produced by our RAG system. Table~\ref{tab:test} shows a part of the test set with 5 questions, generated answers, and the ground truths.

{\customsize
\begin{longtable}{|l|l|l|l|}
\caption{Test Set} 
\label{tab:test} \\
\hline
No & Questions & Answers & Ground Truth \\ 
\hline
\endfirsthead
\hline
No & Questions & Answers & Ground Truth \\ 
\hline
\endhead

\hline
\multicolumn{4}{|r|}{\textit{Continued on next page}} \\
\hline
\endfoot

\hline
\endlastfoot

1 & \begin{tabular}[c]{@{}l@{}}Why AC hi-pot \\ test is used?\end{tabular} & 
\begin{tabular}[c]{@{}l@{}}AC hi-pot test is used \\to assess transformer windings\\ condition. It simulates the \\internal stress that \\transformers face during \\operating conditions.\end{tabular} & 
\begin{tabular}[c]{@{}l@{}}The AC hi-pot test is used to \\ assess transformer windings \\ condition.\end{tabular} \\ 
\hline

2 & \begin{tabular}[c]{@{}l@{}}What should be \\ done to complete \\ the TTR test?\end{tabular} & 
\begin{tabular}[c]{@{}l@{}}The following steps are used for \\completing the TTR test: \\- Transformer is isolated and \\tagged and leads disconnected \\- Check transformer nameplate \\- Check the polarities and vectors \\(phasors) \\- Determine ratios for each \\no-load and load tap position\end{tabular} & 
\begin{tabular}[c]{@{}l@{}}The following steps are used \\ for completing the TTR test: \\ Transformer is isolated and \\ tagged and leads disconnected \\ Check transformer nameplate \\ Check the polarities and vectors \\ (phasors). Determine ratios \\ for each no-load and load tap \\ position\end{tabular} \\ 
\hline

3 & \begin{tabular}[c]{@{}l@{}}A transformer needs \\ to verify the \\ continuity and phase \\ identification of  \\ windings. What tests \\ should be done?\end{tabular} & 
\begin{tabular}[c]{@{}l@{}}The manual describes three \\tests using a 100W lamp \\and a 120V 60Hz power \\supply to verify transformer \\winding continuity and phase \\identification.\\Test 1: Connect the power\\ supply to the primary \\terminals of the transformer \\while leaving the secondary \\open. Observe if the lamp \\lights up dimly.\\Test 2: Short the secondary \\winding while maintaining \\the connections from \\Test 1. The lamp \\should illuminate brightly.\\Test 3: For three-phase \\transformers, repeat Tests 1 \\and 2 for each \\individual winding while \\keeping the other \\windings open.\end{tabular} & 
\begin{tabular}[c]{@{}l@{}}Test 1: Connect the 120 V, \\ 60 Hz power through the \\ lamp to the transformer \\ primary, terminals. Leave \\ the transformer secondary \\ winding open. The lamp \\ will burn dimly. \\ Test 2: Keep connections as \\ presented in test 1, but \\ now short the secondary \\ winding. The lamp should \\ burn with great brilliance. \\ If the lamp still burns with \\ somewhat less than full \\ brilliance, check for issues \\ in the transformer winding.\\ Test 3: This test is similar \\ to tests 1 and 2, but as applied \\ to a three phase transformer \\ for phase identification and \\ phase continuity check. \\ Complete tests 1 and 2 for \\ each winding of a three-phase \\ transformer individually with \\ the remaining windings kept \\ open.\end{tabular} \\ 
\hline

4 & \begin{tabular}[c]{@{}l@{}}Why AC voltage \\ is preferred for \\ transformer testing?\end{tabular} & 
\begin{tabular}[c]{@{}l@{}}AC voltage is \\preferable to DC voltage \\for transformer testing \\because AC voltage \\simulates the internal \\stress that the transformers \\face during operating conditions.\end{tabular} & 
\begin{tabular}[c]{@{}l@{}}AC voltage simulates the \\ internal stress that the \\ transformers face during \\ operating conditions.\end{tabular} \\ 
\hline

5 & \begin{tabular}[c]{@{}l@{}}What are the common \\ internal mechanical \\ issues found in \\ transformers with \\ FRA?\end{tabular} & 
\begin{tabular}[c]{@{}l@{}}- Core displacement \\- Partial winding collapse \\- Faulty core grounds \\- Shorted turns and open windings \\- Broken or loosened clamping \\structures \\- Winding deformation and\\ displacement\end{tabular} & 
\begin{tabular}[c]{@{}l@{}}- Core displacement\\ -Partial winding collapse\\ -Faulty core grounds\\ - Shorted turns and open \\ windings\\ - Broken or loosened clamping \\ structures\\ - Winding deformation and \\ displacement\end{tabular} \\ 
\hline
\end{longtable}}

These 5 questions were selected from a technical document. The other 45 questions were taken from different types of documents, both technical and non-technical ones. We made sure that none of these questions were used in training the system. This helps us get a fair idea of how well the system performs. An example of the system answering general questions from documents like storybooks is shown in Section~\ref{Gen}.

\subsubsection{RAGas Framework:}
RAGas (Retrieval-Augmented Generation assessment) is a framework designed to evaluate RAG systems by measuring key aspects like relevance, faithfulness, and answer completeness \cite{es2024ragas}. In this work, using the RAGas framework, we are measuring \textbf{Faithfulness, Answer Relevancy, Context Precision, Context Recall}- these 4 metrics. 

\begin{itemize}
    \item \textbf{Faithfulness:} Measures how accurately the generated response aligns with the retrieved context or factual correctness of the answer.
    \item \textbf{Answer Relevancy:} Evaluates how well the generated answer addresses the user’s query or precise answers.
    \item \textbf{Context Precision:} Assesses the proportion of retrieved context that is relevant to generating a correct and informative response.
    \item \textbf{Context Recall:} Measures how much of the necessary and relevant information from the available context was successfully retrieved to support the response.
\end{itemize}

Our evaluated results are given in Table \ref{tab:RAGas}:

\begin{table}[h]
\caption{RAGas Evaluation Score}
\label{tab:RAGas}
\resizebox{\textwidth}{!}{%
\begin{tabular}{|l|l|l|l|l|}
\hline
\begin{tabular}[c]{@{}l@{}}Question \\ No\end{tabular} & Faithfulness & Answer Relevancy & Context Precision & Context Recall \\ \hline
1             & 0.5 & 0.748477   & 1 & 1 \\ \hline
2             & 1   & 0.933533   & 1 & 1 \\ \hline
3             & 1   & 0.963467   & 1 & 1 \\ \hline
4             & 1   & 0.964333   & 1 & 1 \\ \hline
5             & 1   & 0.687372   & 1 & 1 \\ \hline
% \begin{tabular}[c]{@{}l@{}}Average Score\\ of 5 questions\end{tabular}  & 0.9          & 0.85944055       & 1                 & 1              \\ \hline
\begin{tabular}[c]{@{}l@{}}Average Score\\ of 35 questions\\ from Technical\\ Documents\end{tabular} & 0.9357         & 0.86156233       & 0.9429              & 0.9714           \\ \hline

\begin{tabular}[c]{@{}l@{}}Average Score\\ of 15 questions\\ from Non-Technical\\ Documents\end{tabular} & 0.95         & 0.89225678       & 0.9333              & 0.967           \\ \hline

\begin{tabular}[c]{@{}l@{}}Average Score\\ of 50 questions\end{tabular} & 0.94         & 0.87077067       & 0.94              & 0.97           \\ \hline
\end{tabular}}
\end{table}

From the scores, it is clear that both context precision and context recall are excellent. RAGas used the same chunk size as the proposed pipeline, and so the correct context and retrieved context are similar in size, leading Precision and Recall to very high value if the retrieved context is correct. Faithfulness is also very good, indicating the answer is factually correct. Answer relevancy is also good, but not perfect. 

The reason is that for the answer relevancy to be perfect, no extra words compared to the ground truth should be present. But it is generally not possible. Thus, the answer relevancy is never 1. Scores are generally higher on simple or non-technical documents, because they do not carry complex tables/ images/ information like technical documents. 

\subsubsection{DeepEval Framework:}
The metrics we used using DeepEval are \textbf{Faithfulness, Answer Relevancy, and Contextual Relevancy.}

% Contextual Relevancy evaluates the extent to which the response aligns with the context provided, ensuring that the generated text maintains coherence and stays on topic with the surrounding information. Contextual Relevancy is different than Context Precision and Context Recall.

Contextual Relevancy focuses on the overall relevance of the answer to the context. Even though both Contextual Relevancy and Faithfulness talk about context and answer, they are different. Faithfulness measures factual correctness, whereas Contextual Relevancy measures topical alignment of the answer with the retrieved context.

Evaluated results by DeepEval are given in Table \ref{tab:Deepeval}:

\begin{table}[h]
\centering
\caption{DeepEval Evaluation Score}
\label{tab:Deepeval}
\begin{tabular}{|l|l|l|l|}
\hline
\begin{tabular}[c]{@{}l@{}}Question \\ No\end{tabular} & Faithfulness & Answer Relevancy & Contextual Relevancy \\ \hline
1             & 1    & 1 & 0.9    \\ \hline
2             & 1    & 1 & 0.9333 \\ \hline
3             & 1    & 1 & 1      \\ \hline
4             & 0.75 & 1 & 1      \\ \hline
5             & 1    & 1 & 0.5862 \\ \hline
% \begin{tabular}[c]{@{}l@{}}Average Score\\ of 5 questions\end{tabular}  & 0.95          & 1       & 0.8839  \\ \hline
\begin{tabular}[c]{@{}l@{}}Average Score\\ of 35 questions\\ from Technical\\ Documents\end{tabular}  & 0.9571          & 0.9286       & 0.8839  \\ \hline
\begin{tabular}[c]{@{}l@{}}Average Score\\ of 15 questions\\ from Non-Technical\\ Documents\end{tabular}  & 0.967          & 0.9333       & 0.9122  \\ \hline
\begin{tabular}[c]{@{}l@{}}Average Score\\ of 50 questions\end{tabular} & 0.96         & 0.93       & 0.89239           \\ \hline
\end{tabular}
\end{table}

DeepEval explains the reasoning for the score.

\textbf{Metrics Summary of Question 1 from DeepEval is as following:}

\begin{itemize}
  \item \textbf{Answer Relevancy} (score: 1.0, threshold: 0.5, strict: False, reason: The score is 1.00 because the output directly and accurately addresses the user's question about the purpose of an AC hi-pot test.., error: None)
  \item \textbf{Contextual Relevancy} (score: 0.9, threshold: 0.6, strict: False, reason: The score is 0.90 because while the retrieval context discusses the AC hi-pot test, it also includes information about the TTR test, which is irrelevant to the input. For example, it states 'This statement is about the TTR test and not the AC hi-pot test.' However, the context does provide relevant details about the AC hi-pot test, such as 'The AC hi-pot test is used to assess transformer windings condition.', error: None)
  \item \textbf{Faithfulness} (score: 1.0, threshold: 0.7, strict: False, reason: The output is perfectly aligned with the context! Great job!, error: None)
\end{itemize}

The 2 frameworks sometimes show different scores because RAGas evaluates exactly word by word, and DeepEval measures according to meaning. So the score is lower in RAGas compared to DeepEval, as exact words have to be present for a higher score in RAGas.

\subsection{Comparison between General vs Proposed RAG Pipeline:}
\label{ABLATION}
\begin{figure}[h]
    \centering
    \includegraphics[width=0.8\textwidth]{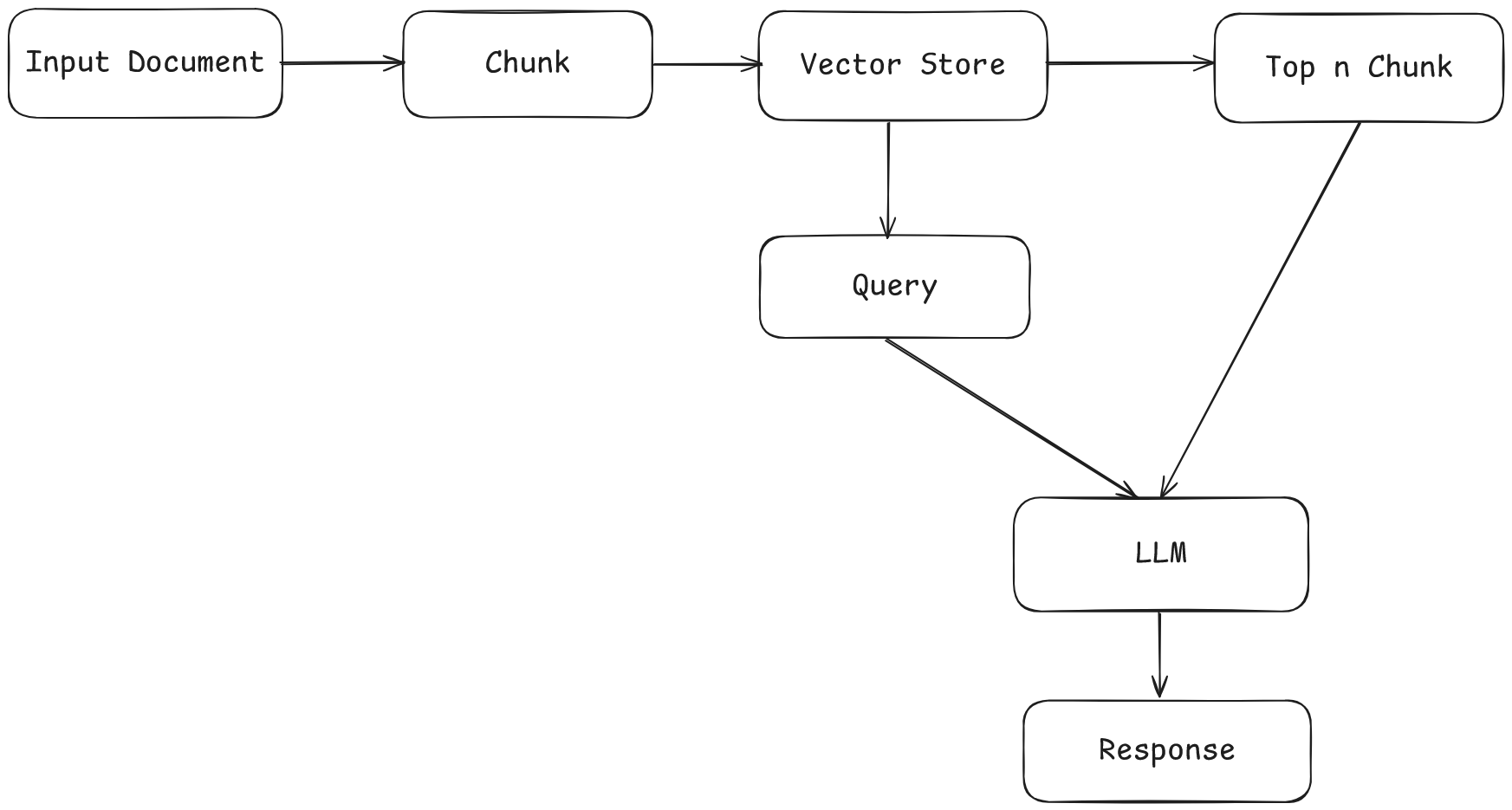}
    \caption{General RAG Pipeline}
    \label{fig:GRAG}
\end{figure}

The architecture of General RAG pipelines consists of single-step retrieval with generally no processing except chunking and conversion to embeddings. The typical RAG architecture is given in Figure \ref{fig:GRAG}.

Here we will compare the generated response of our proposed RAG pipeline with typical RAG pipelines. For exact comparison, we have used the same embeddings and vector stores as in the proposed pipeline.

Table \ref{tab:ABRES} shows the response of both pipelines.

{\customsize
\begin{longtable}{|l|l|l|l|}
\caption{Comparison between Proposed Pipeline \& General Pipeline } \label{tab:ABRES} \\
\hline
No & Questions & Proposed Pipeline & General Pipeline \\ 
\hline
\endfirsthead
\hline
No & Questions & Proposed Pipeline & General Pipeline \\ 
\hline
\endhead

\hline
\multicolumn{4}{|r|}{\textit{Continued on next page}} \\
\hline
\endfoot

\hline
\endlastfoot

1 & \begin{tabular}[c]{@{}l@{}}For what gases \\transformer assets are \\recoverable?\end{tabular} & 
\begin{tabular}[c]{@{}l@{}}The "C2H2 and H2" (gases 2) \\combination is suitable for \\recoverable Smaller distribution \\transformers.\end{tabular} & 
\begin{tabular}[c]{@{}l@{}}According to the text, \\"Recoverable" transformers typically \\use online DGA tools \\that measure only **H2** \\(Hydrogen).\end{tabular} \\ 
\hline

2 & \begin{tabular}[c]{@{}l@{}}What should be \\ done to complete the \\ TTR test?\end{tabular} & 
\begin{tabular}[c]{@{}l@{}}The following steps are used for \\completing the TTR test: \\- Transformer is isolated and \\tagged and leads disconnected \\- Check transformer nameplate \\- Check the polarities and vectors \\(phasors) \\- Determine ratios for each \\no-load and load tap position\end{tabular} & 
\begin{tabular}[c]{@{}l@{}}Here are the steps involved \\in completing a TTR test\\-Isolate and Tag: Disconnect \\the transformer from its \\power source and tag it \\to indicate it is out of service.\\Check Nameplate: Review the \\transformer's nameplate for \\information about its \\rated voltage, turns ratio, \\and polarity.\\Determine Ratios: Using the \\TTR test set, apply voltage \\to one winding and measure \\the voltage induced in \\the other winding. \\This will give you the turns ratio.\\Verify Polarity: Ensure \\the polarity of the \\transformer windings matches \\what is indicated on the nameplate.\\ position\end{tabular} \\ 
\hline

3 & \begin{tabular}[c]{@{}l@{}}A transformer needs \\ to verify the continuity \\ and phase identification \\ of windings. What tests \\ should be done?\end{tabular} & 
\begin{tabular}[c]{@{}l@{}}The manual describes three \\tests using a 100W lamp \\and a 120V 60Hz power \\supply to verify transformer \\winding continuity and phase \\identification.\\Test 1: Connect the power\\ supply to the primary \\terminals of the transformer \\while leaving the secondary \\open. Observe if the lamp \\lights up dimly.\\Test 2: Short the secondary \\winding while maintaining \\the connections from \\Test 1. The lamp \\should illuminate brightly.\\Test 3: For three-phase \\transformers, repeat Tests 1 \\and 2 for each \\individual winding while \\keeping the other \\windings open.\end{tabular} & 
\begin{tabular}[c]{@{}l@{}}A fast and rough test can be \\completed to verify the continuity \\and phase identification of \\transformer windings. This test uses \\a 100W lamp with a socket \\and an extension cord connected \\to a 120V 60Hz power supply. \\Test procedures are outlined:\\Test 1: Connect the \\power supply to the primary \\terminals while leaving the \\secondary winding open. The \\lamp should burn dimly.\\Test 2: Short the secondary winding \\while keeping the connections \\from Test 1. The lamp should burn \\brightly. If it doesn't, there might \\be an issue with the transformer \\winding. \\Test 3: Repeat Tests 1 and 2 for \\each winding of a three-phase \\transformer individually, keeping \\the other windings open.\end{tabular} \\ 
\hline

4 & \begin{tabular}[c]{@{}l@{}}Why AC voltage \\ is preferred for \\ transformer testing?\end{tabular} & 
\begin{tabular}[c]{@{}l@{}}AC voltage is \\preferable to DC voltage \\for transformer testing \\because AC voltage \\simulates the internal \\stress that the transformers \\face during operating conditions.\end{tabular} & 
\begin{tabular}[c]{@{}l@{}}AC voltage is preferred \\for transformer testing because \\it simulates the internal stresses \\that transformers experience \\during normal operation.\end{tabular} \\ 
\hline

5 & \begin{tabular}[c]{@{}l@{}}What should be done \\before applying rated \\service voltage to a\\ transformer? Answer \\from given document.\end{tabular} & 
\begin{tabular}[c]{@{}l@{}}This document doesn't \\ contain the answer\end{tabular} & 
\begin{tabular}[c]{@{}l@{}}The voltage needs to be started \\at one-quarter or less of the full \\value and increased up \\to full value in not \\more than 15 seconds.\end{tabular} \\ 
\hline
\end{longtable}}

The quantitative performance of our proposed model with typical RAG systems is given in Table \ref{tab:Ablation}. The performance was measured using the RAGas framework using \textbf{Llama-3.3-70b LLM}.

\begin{table}[htbp]
\centering
\caption{Comparison between Typical \& Proposed Pipeline}
\label{tab:Ablation}
\begin{NiceTabular}{|l|l|l|l|l|l|}
\hline
Question No & Pipeline & Faithfulness & \makecell{Answer \\ Relevancy} & \makecell{Context \\ Precision} & \makecell{Context \\ Recall} \\ \hline

\multirow{2}{*}{Q1} & Proposed & 0.5     & 0.710086  & 1   & 1   \\ \cline{2-6} 
                    & General  & 0       & 0.576561  & 1   & 1   \\ \hline

\multirow{2}{*}{Q2} & Proposed & 1       & 0.933533  & 1   & 1   \\ \cline{2-6} 
                    & General  & 0.7778  & 0.826518  & 1   & 1   \\ \hline

\multirow{2}{*}{Q3} & Proposed & 1       & 0.963467  & 1   & 1   \\ \cline{2-6} 
                    & General  & 1       & 0.939695  & 1   & 1   \\ \hline

\multirow{2}{*}{Q4} & Proposed & 1       & 0.964333  & 1   & 1   \\ \cline{2-6} 
                    & General  & 1       & 0.996398  & 1   & 1   \\ \hline

\multirow{2}{*}{Q5} & Proposed & 0.8322  & 0.92765   & N/A   & N/A   \\ \cline{2-6} 
                    & General  & 0       & 0.4301    & N/A   & N/A   \\ \hline

\multirow{2}{*}{Average} & Proposed & 0.86644 & 0.8998138 & 1 & 1 \\ \cline{2-6} 
                         & General  & 0.55556 & 0.7538544 & 1 & 1 \\ \hline

\end{NiceTabular}
\end{table}

The question \textbf{Q1} was set from a Table. As we can see, the general pipelines could not answer properly, whereas the proposed pipeline successfully answered. This proved the capability of the \textbf{table data extraction} of our pipeline.

\textbf{Q2 to Q4} were general questions. Both architectures answered it successfully.

\textbf{Q5} was set from outside the document. As there was no correct context, context recall and context precision would always be 0, no matter what. And so, these two metrics were skipped. The proposed pipeline correctly identified that the answer was not present in the document. On the other hand, the typical RAG pipeline suffered from hallucination because the question was about transformer, and also the given document was about transformer, even though it did not carry the answer. But using vector search to retrieve context might bring context that may not carry the correct answer. The reason the proposed architecture performed better in this case is because of \textbf{RAFT}, the fine-tuning approach that helped LLM to distinguish correct and false context. As the retrieved context is false, LLM did not answer.

\subsection{Performance on General Document Question-Answering:}
\label{Gen}
From earlier evaluations, it is proven that our model performs better in all sorts of document-based question answering. Here, the generated answers on questions from some general documents will be discussed.

For this comparison, we used a scanned document. The document can be checked on from \href{https://drive.google.com/file/d/1FYN2x35kwgzjJTkFo763o1UCZ_aAz7Hg/view?usp=sharing}{here}. Using the proposed methodology, we converted it to a searchable PDF which can be accessed from \href{https://drive.google.com/file/d/1CO0jRsJ0eOH-EGk5F3V9OKPPygyc7gxO/view?usp=sharing}{here}.

This is not a technical manual. So the evaluation will help us understand the proposed pipeline's subjective performance on a \textbf{general scanned document}. The questions and answers are given in \ref{tab:Per}.
%%%%%%%%%%%%%%%%%%%%%%%%%%%%%%%%%%%%%%%%%%%%%%%%%%%%%%%%%%%%%%%%%%%%%%%%%%%%%%%%%
\begin{table}
\centering
\caption{Performance on General Scanned Document}
\label{tab:Per}
\resizebox{\textwidth}{!}{%
\begin{tabular}{|l|l|} 
\hline
Questions                                                                                                             & Answers                                                                                                                                                                                                                                                                                                                                                                                                     \\ 
\hline
\begin{tabular}[c]{@{}l@{}}Who had powerful spirit \\obedient to his will?\end{tabular}                               & Prospero                                                                                                                                                                                                                                                                                                                                                                                                    \\ 
\hline
\begin{tabular}[c]{@{}l@{}}Why did the duke went \\with a few faithful \\followers? And where did he go?\end{tabular} & \begin{tabular}[c]{@{}l@{}}The duke went with a few faithful followers because he had\\~been driven from his dukedom by his younger brother, who \\had taken it unlawfully. He went to the forest of Arden.\end{tabular}                                                                                                                                                                                    \\ 
\hline
Why Celia went with Rosalind                                                                                          & \begin{tabular}[c]{@{}l@{}}Celia went with Rosalind because she couldn't persuade her \\father to let Rosalind remain with her.\end{tabular}                                                                                                                                                                                                                                                                \\ 
\hline
\begin{tabular}[c]{@{}l@{}}Why did Macbeth kill \\the king?\end{tabular}                                              & \begin{tabular}[c]{@{}l@{}}The text states that Macbeth killed King Duncan because the \\witches prophesied that he would become king.\end{tabular}                                                                                                                                                                                                                                                         \\ 
\hline
\begin{tabular}[c]{@{}l@{}}Explain the special \\law in Athens\end{tabular}                                           & \begin{tabular}[c]{@{}l@{}}In Athens, there was a law that allowed fathers to force their \\daughters to marry whomever they chose. If a daughter \\disobeyed her father and refused to marry the chosen \\man, the father could demand the death penalty be carried \\out against her. However, this law was rarely enforced as \\fathers typically didn't want to kill their own daughters.\end{tabular}  \\
\hline
\end{tabular}}
\end{table}
%%%%%%%%%%%%%%%%%%%%%%%%%%%%%%%%%%%%%%%%%%%%%%%%%%%%%%
From the table, it is clear that our proposed methodology works well on all types of documents, whether it is in scanned or searchable format.

\section{Conclusion}\label{sec6}
This work introduced an enhanced Retrieval-Augmented Generation (RAG) pipeline designed to improve information extraction from both structured and unstructured data. By combining dual-stage retrieval, semantic search, and a fine-tuned reranker using the RAFT (Retrieval-Augmented Fine-Tuning) method, the system achieved more accurate, context-aware responses. Through our methodology, accurate answers can be generated from both technical and non-technical documents.

Even though our pipeline has been proven to be superior to a general RAG pipeline, there are still some issues. Currently, information extraction from flowcharts remains an open challenge. So, it could not be incorporated in this work. 
We tried to implement RAG with all open-source models. But the limited performance of open-source Vision-Language Models (VLMs) presents challenges in accurate image-based information extraction.

Future improvements include improving VLM capabilities for better visual understanding and developing techniques to extract structured information from flowcharts. These enhancements would significantly improve the pipeline’s ability to handle multimodal data and expand its effectiveness in real-world applications.

\backmatter
\bmhead{Supplementary information}
For reproducibility, all the notebooks and materials are available in \href{https://github.com/ShadmanSobhan/SDA-RAG}{GitHub}.

%\bmhead{Acknowledgements}

\bibliography{sn-bibliography}


\begin{thebibliography}{42}
\ifx \bisbn   \undefined \def \bisbn  #1{ISBN #1}\fi
\ifx \binits  \undefined \def \binits#1{#1}\fi
\ifx \bauthor  \undefined \def \bauthor#1{#1}\fi
\ifx \batitle  \undefined \def \batitle#1{#1}\fi
\ifx \bjtitle  \undefined \def \bjtitle#1{#1}\fi
\ifx \bvolume  \undefined \def \bvolume#1{\textbf{#1}}\fi
\ifx \byear  \undefined \def \byear#1{#1}\fi
\ifx \bissue  \undefined \def \bissue#1{#1}\fi
\ifx \bfpage  \undefined \def \bfpage#1{#1}\fi
\ifx \blpage  \undefined \def \blpage #1{#1}\fi
\ifx \burl  \undefined \def \burl#1{\textsf{#1}}\fi
\ifx \doiurl  \undefined \def \doiurl#1{\url{https://doi.org/#1}}\fi
\ifx \betal  \undefined \def \betal{\textit{et al.}}\fi
\ifx \binstitute  \undefined \def \binstitute#1{#1}\fi
\ifx \binstitutionaled  \undefined \def \binstitutionaled#1{#1}\fi
\ifx \bctitle  \undefined \def \bctitle#1{#1}\fi
\ifx \beditor  \undefined \def \beditor#1{#1}\fi
\ifx \bpublisher  \undefined \def \bpublisher#1{#1}\fi
\ifx \bbtitle  \undefined \def \bbtitle#1{#1}\fi
\ifx \bedition  \undefined \def \bedition#1{#1}\fi
\ifx \bseriesno  \undefined \def \bseriesno#1{#1}\fi
\ifx \blocation  \undefined \def \blocation#1{#1}\fi
\ifx \bsertitle  \undefined \def \bsertitle#1{#1}\fi
\ifx \bsnm \undefined \def \bsnm#1{#1}\fi
\ifx \bsuffix \undefined \def \bsuffix#1{#1}\fi
\ifx \bparticle \undefined \def \bparticle#1{#1}\fi
\ifx \barticle \undefined \def \barticle#1{#1}\fi
\bibcommenthead
\ifx \bconfdate \undefined \def \bconfdate #1{#1}\fi
\ifx \botherref \undefined \def \botherref #1{#1}\fi
\ifx \url \undefined \def \url#1{\textsf{#1}}\fi
\ifx \bchapter \undefined \def \bchapter#1{#1}\fi
\ifx \bbook \undefined \def \bbook#1{#1}\fi
\ifx \bcomment \undefined \def \bcomment#1{#1}\fi
\ifx \oauthor \undefined \def \oauthor#1{#1}\fi
\ifx \citeauthoryear \undefined \def \citeauthoryear#1{#1}\fi
\ifx \endbibitem  \undefined \def \endbibitem {}\fi
\ifx \bconflocation  \undefined \def \bconflocation#1{#1}\fi
\ifx \arxivurl  \undefined \def \arxivurl#1{\textsf{#1}}\fi
\csname PreBibitemsHook\endcsname

%%% 1
\bibitem[\protect\citeauthoryear{Vaswani}{2017}]{vaswani2017attention}
\begin{botherref}
\oauthor{\bsnm{Vaswani}, \binits{A.}}:
Attention is all you need.
Advances in Neural Information Processing Systems
(2017)
\end{botherref}
\endbibitem

%%% 2
\bibitem[\protect\citeauthoryear{Mousavi et~al.}{2024}]{mousavi2024your}
\begin{botherref}
\oauthor{\bsnm{Mousavi}, \binits{S.M.}},
\oauthor{\bsnm{Alghisi}, \binits{S.}},
\oauthor{\bsnm{Riccardi}, \binits{G.}}:
Is your llm outdated? benchmarking llms \& alignment algorithms for time-sensitive knowledge.
arXiv preprint arXiv:2404.08700
(2024)
\end{botherref}
\endbibitem

%%% 3
\bibitem[\protect\citeauthoryear{Reddy et~al.}{2024}]{reddy2024hallucinations}
\begin{bchapter}
\bauthor{\bsnm{Reddy}, \binits{G.P.}},
\bauthor{\bsnm{Kumar}, \binits{Y.P.}},
\bauthor{\bsnm{Prakash}, \binits{K.P.}}:
\bctitle{Hallucinations in large language models (llms)}.
In: \bbtitle{2024 IEEE Open Conference of Electrical, Electronic and Information Sciences (eStream)},
pp. \bfpage{1}--\blpage{6}
(\byear{2024}).
\bcomment{IEEE}
\end{bchapter}
\endbibitem

%%% 4
\bibitem[\protect\citeauthoryear{Banerjee et~al.}{2024}]{banerjee2024llms}
\begin{botherref}
\oauthor{\bsnm{Banerjee}, \binits{S.}},
\oauthor{\bsnm{Agarwal}, \binits{A.}},
\oauthor{\bsnm{Singla}, \binits{S.}}:
Llms will always hallucinate, and we need to live with this.
arXiv preprint arXiv:2409.05746
(2024)
\end{botherref}
\endbibitem

%%% 5
\bibitem[\protect\citeauthoryear{Augenstein et~al.}{2024}]{augenstein2024factuality}
\begin{barticle}
\bauthor{\bsnm{Augenstein}, \binits{I.}},
\bauthor{\bsnm{Baldwin}, \binits{T.}},
\bauthor{\bsnm{Cha}, \binits{M.}},
\bauthor{\bsnm{Chakraborty}, \binits{T.}},
\bauthor{\bsnm{Ciampaglia}, \binits{G.L.}},
\bauthor{\bsnm{Corney}, \binits{D.}},
\bauthor{\bsnm{DiResta}, \binits{R.}},
\bauthor{\bsnm{Ferrara}, \binits{E.}},
\bauthor{\bsnm{Hale}, \binits{S.}},
\bauthor{\bsnm{Halevy}, \binits{A.}}, \betal:
\batitle{Factuality challenges in the era of large language models and opportunities for fact-checking}.
\bjtitle{Nature Machine Intelligence}
\bvolume{6}(\bissue{8}),
\bfpage{852}--\blpage{863}
(\byear{2024})
\end{barticle}
\endbibitem

%%% 6
\bibitem[\protect\citeauthoryear{Zhuang et~al.}{2023}]{zhuang2023toolqa}
\begin{barticle}
\bauthor{\bsnm{Zhuang}, \binits{Y.}},
\bauthor{\bsnm{Yu}, \binits{Y.}},
\bauthor{\bsnm{Wang}, \binits{K.}},
\bauthor{\bsnm{Sun}, \binits{H.}},
\bauthor{\bsnm{Zhang}, \binits{C.}}:
\batitle{Toolqa: A dataset for llm question answering with external tools}.
\bjtitle{Advances in Neural Information Processing Systems}
\bvolume{36},
\bfpage{50117}--\blpage{50143}
(\byear{2023})
\end{barticle}
\endbibitem

%%% 7
\bibitem[\protect\citeauthoryear{Huang et~al.}{2023}]{huang2023dsqa}
\begin{bchapter}
\bauthor{\bsnm{Huang}, \binits{D.}},
\bauthor{\bsnm{Wei}, \binits{Z.}},
\bauthor{\bsnm{Yue}, \binits{A.}},
\bauthor{\bsnm{Zhao}, \binits{X.}},
\bauthor{\bsnm{Chen}, \binits{Z.}},
\bauthor{\bsnm{Li}, \binits{R.}},
\bauthor{\bsnm{Jiang}, \binits{K.}},
\bauthor{\bsnm{Chang}, \binits{B.}},
\bauthor{\bsnm{Zhang}, \binits{Q.}},
\bauthor{\bsnm{Zhang}, \binits{S.}}, \betal:
\bctitle{Dsqa-llm: domain-specific intelligent question answering based on large language model}.
In: \bbtitle{International Conference on AI-generated Content},
pp. \bfpage{170}--\blpage{180}
(\byear{2023}).
\bcomment{Springer}
\end{bchapter}
\endbibitem

%%% 8
\bibitem[\protect\citeauthoryear{Mohammad et~al.}{2023}]{mohammad2023large}
\begin{bchapter}
\bauthor{\bsnm{Mohammad}, \binits{A.F.}},
\bauthor{\bsnm{Clark}, \binits{B.}},
\bauthor{\bsnm{Hegde}, \binits{R.}}:
\bctitle{Large language model (llm) \& gpt, a monolithic study in generative ai}.
In: \bbtitle{2023 Congress in Computer Science, Computer Engineering, \& Applied Computing (CSCE)},
pp. \bfpage{383}--\blpage{388}
(\byear{2023}).
\bcomment{IEEE}
\end{bchapter}
\endbibitem

%%% 9
\bibitem[\protect\citeauthoryear{Gaikwad et~al.}{2022}]{gaikwad2022extensive}
\begin{botherref}
\oauthor{\bsnm{Gaikwad}, \binits{A.}},
\oauthor{\bsnm{Rambhia}, \binits{P.}},
\oauthor{\bsnm{Pawar}, \binits{S.}}:
An extensive analysis between different language models: gpt-3, bert and macaw
(2022)
\end{botherref}
\endbibitem

%%% 10
\bibitem[\protect\citeauthoryear{Devlin et~al.}{2019}]{devlin2019bert}
\begin{bchapter}
\bauthor{\bsnm{Devlin}, \binits{J.}},
\bauthor{\bsnm{Chang}, \binits{M.-W.}},
\bauthor{\bsnm{Lee}, \binits{K.}},
\bauthor{\bsnm{Toutanova}, \binits{K.}}:
\bctitle{Bert: Pre-training of deep bidirectional transformers for language understanding}.
In: \bbtitle{Proceedings of the 2019 Conference of the North American Chapter of the Association for Computational Linguistics: Human Language Technologies, Volume 1 (long and Short Papers)},
pp. \bfpage{4171}--\blpage{4186}
(\byear{2019})
\end{bchapter}
\endbibitem

%%% 11
\bibitem[\protect\citeauthoryear{Artetxe et~al.}{2022}]{artetxe2022role}
\begin{botherref}
\oauthor{\bsnm{Artetxe}, \binits{M.}},
\oauthor{\bsnm{Du}, \binits{J.}},
\oauthor{\bsnm{Goyal}, \binits{N.}},
\oauthor{\bsnm{Zettlemoyer}, \binits{L.}},
\oauthor{\bsnm{Stoyanov}, \binits{V.}}:
On the role of bidirectionality in language model pre-training.
arXiv preprint arXiv:2205.11726
(2022)
\end{botherref}
\endbibitem

%%% 12
\bibitem[\protect\citeauthoryear{{\"O}z{\c{c}}ift et~al.}{2021}]{ozccift2021advancing}
\begin{barticle}
\bauthor{\bsnm{{\"O}z{\c{c}}ift}, \binits{A.}},
\bauthor{\bsnm{Akarsu}, \binits{K.}},
\bauthor{\bsnm{Yumuk}, \binits{F.}},
\bauthor{\bsnm{S{\"o}ylemez}, \binits{C.}}:
\batitle{Advancing natural language processing (nlp) applications of morphologically rich languages with bidirectional encoder representations from transformers (bert): an empirical case study for turkish}.
\bjtitle{Automatika: {\v{c}}asopis za automatiku, mjerenje, elektroniku, ra{\v{c}}unarstvo i komunikacije}
\bvolume{62}(\bissue{2}),
\bfpage{226}--\blpage{238}
(\byear{2021})
\end{barticle}
\endbibitem

%%% 13
\bibitem[\protect\citeauthoryear{Wu et~al.}{2022}]{wu2022study}
\begin{barticle}
\bauthor{\bsnm{Wu}, \binits{X.}},
\bauthor{\bsnm{Xia}, \binits{Y.}},
\bauthor{\bsnm{Zhu}, \binits{J.}},
\bauthor{\bsnm{Wu}, \binits{L.}},
\bauthor{\bsnm{Xie}, \binits{S.}},
\bauthor{\bsnm{Qin}, \binits{T.}}:
\batitle{A study of bert for context-aware neural machine translation}.
\bjtitle{Machine Learning}
\bvolume{111}(\bissue{3}),
\bfpage{917}--\blpage{935}
(\byear{2022})
\end{barticle}
\endbibitem

%%% 14
\bibitem[\protect\citeauthoryear{Radford et~al.}{2018}]{radford2018improving}
\begin{botherref}
\oauthor{\bsnm{Radford}, \binits{A.}},
\oauthor{\bsnm{Narasimhan}, \binits{K.}},
\oauthor{\bsnm{Salimans}, \binits{T.}},
\oauthor{\bsnm{Sutskever}, \binits{I.}}, et al.:
Improving language understanding by generative pre-training
(2018)
\end{botherref}
\endbibitem

%%% 15
\bibitem[\protect\citeauthoryear{Radford et~al.}{2019}]{radford2019language}
\begin{barticle}
\bauthor{\bsnm{Radford}, \binits{A.}},
\bauthor{\bsnm{Wu}, \binits{J.}},
\bauthor{\bsnm{Child}, \binits{R.}},
\bauthor{\bsnm{Luan}, \binits{D.}},
\bauthor{\bsnm{Amodei}, \binits{D.}},
\bauthor{\bsnm{Sutskever}, \binits{I.}}, \betal:
\batitle{Language models are unsupervised multitask learners}.
\bjtitle{OpenAI blog}
\bvolume{1}(\bissue{8}),
\bfpage{9}
(\byear{2019})
\end{barticle}
\endbibitem

%%% 16
\bibitem[\protect\citeauthoryear{Brown et~al.}{2020}]{brown2020language}
\begin{barticle}
\bauthor{\bsnm{Brown}, \binits{T.}},
\bauthor{\bsnm{Mann}, \binits{B.}},
\bauthor{\bsnm{Ryder}, \binits{N.}},
\bauthor{\bsnm{Subbiah}, \binits{M.}},
\bauthor{\bsnm{Kaplan}, \binits{J.D.}},
\bauthor{\bsnm{Dhariwal}, \binits{P.}},
\bauthor{\bsnm{Neelakantan}, \binits{A.}},
\bauthor{\bsnm{Shyam}, \binits{P.}},
\bauthor{\bsnm{Sastry}, \binits{G.}},
\bauthor{\bsnm{Askell}, \binits{A.}}, \betal:
\batitle{Language models are few-shot learners}.
\bjtitle{Advances in neural information processing systems}
\bvolume{33},
\bfpage{1877}--\blpage{1901}
(\byear{2020})
\end{barticle}
\endbibitem

%%% 17
\bibitem[\protect\citeauthoryear{Gao et~al.}{2023}]{gao2023examining}
\begin{botherref}
\oauthor{\bsnm{Gao}, \binits{K.}},
\oauthor{\bsnm{He}, \binits{S.}},
\oauthor{\bsnm{He}, \binits{Z.}},
\oauthor{\bsnm{Lin}, \binits{J.}},
\oauthor{\bsnm{Pei}, \binits{Q.}},
\oauthor{\bsnm{Shao}, \binits{J.}},
\oauthor{\bsnm{Zhang}, \binits{W.}}:
Examining user-friendly and open-sourced large gpt models: A survey on language, multimodal, and scientific gpt models.
arXiv preprint arXiv:2308.14149
(2023)
\end{botherref}
\endbibitem

%%% 18
\bibitem[\protect\citeauthoryear{Chen et~al.}{2024}]{chen2024systematic}
\begin{bchapter}
\bauthor{\bsnm{Chen}, \binits{C.}},
\bauthor{\bsnm{Wang}, \binits{B.}},
\bauthor{\bsnm{Lin}, \binits{Y.}}:
\bctitle{A systematic mapping study of llm applications in mobile device research}.
In: \bbtitle{Asia-Pacific Web (APWeb) and Web-Age Information Management (WAIM) Joint International Conference on Web and Big Data},
pp. \bfpage{163}--\blpage{174}
(\byear{2024}).
\bcomment{Springer}
\end{bchapter}
\endbibitem

%%% 19
\bibitem[\protect\citeauthoryear{Chkirbene et~al.}{2024}]{chkirbene2024large}
\begin{bchapter}
\bauthor{\bsnm{Chkirbene}, \binits{Z.}},
\bauthor{\bsnm{Hamila}, \binits{R.}},
\bauthor{\bsnm{Gouissem}, \binits{A.}},
\bauthor{\bsnm{Devrim}, \binits{U.}}:
\bctitle{Large language models (llm) in industry: A survey of applications, challenges, and trends}.
In: \bbtitle{2024 IEEE 21st International Conference on Smart Communities: Improving Quality of Life Using AI, Robotics and IoT (HONET)},
pp. \bfpage{229}--\blpage{234}
(\byear{2024}).
\bcomment{IEEE}
\end{bchapter}
\endbibitem

%%% 20
\bibitem[\protect\citeauthoryear{Yunianto et~al.}{2020}]{yunianto2020domain}
\begin{bchapter}
\bauthor{\bsnm{Yunianto}, \binits{I.}},
\bauthor{\bsnm{Permanasari}, \binits{A.E.}},
\bauthor{\bsnm{Widyawan}, \binits{W.}}:
\bctitle{Domain-specific contextualized embedding: a systematic literature review}.
In: \bbtitle{2020 12th International Conference on Information Technology and Electrical Engineering (ICITEE)},
pp. \bfpage{162}--\blpage{167}
(\byear{2020}).
\bcomment{IEEE}
\end{bchapter}
\endbibitem

%%% 21
\bibitem[\protect\citeauthoryear{Zhang and Li}{2021}]{zhang2021commentary}
\begin{barticle}
\bauthor{\bsnm{Zhang}, \binits{M.}},
\bauthor{\bsnm{Li}, \binits{J.}}:
\batitle{A commentary of gpt-3 in mit technology review 2021}.
\bjtitle{Fundamental Research}
\bvolume{1}(\bissue{6}),
\bfpage{831}--\blpage{833}
(\byear{2021})
\end{barticle}
\endbibitem

%%% 22
\bibitem[\protect\citeauthoryear{Matsui et~al.}{2024}]{matsui2024human}
\begin{barticle}
\bauthor{\bsnm{Matsui}, \binits{K.}},
\bauthor{\bsnm{Utsumi}, \binits{T.}},
\bauthor{\bsnm{Aoki}, \binits{Y.}},
\bauthor{\bsnm{Maruki}, \binits{T.}},
\bauthor{\bsnm{Takeshima}, \binits{M.}},
\bauthor{\bsnm{Takaesu}, \binits{Y.}}:
\batitle{Human-comparable sensitivity of large language models in identifying eligible studies through title and abstract screening: 3-layer strategy using gpt-3.5 and gpt-4 for systematic reviews}.
\bjtitle{Journal of Medical Internet Research}
\bvolume{26},
\bfpage{52758}
(\byear{2024})
\end{barticle}
\endbibitem

%%% 23
\bibitem[\protect\citeauthoryear{Zoph and Le}{2016}]{zoph2016neural}
\begin{botherref}
\oauthor{\bsnm{Zoph}, \binits{B.}},
\oauthor{\bsnm{Le}, \binits{Q.V.}}:
Neural architecture search with reinforcement learning.
arXiv preprint arXiv:1611.01578
(2016)
\end{botherref}
\endbibitem

%%% 24
\bibitem[\protect\citeauthoryear{Dhingra et~al.}{2016}]{dhingra2016gated}
\begin{botherref}
\oauthor{\bsnm{Dhingra}, \binits{B.}},
\oauthor{\bsnm{Liu}, \binits{H.}},
\oauthor{\bsnm{Yang}, \binits{Z.}},
\oauthor{\bsnm{Cohen}, \binits{W.W.}},
\oauthor{\bsnm{Salakhutdinov}, \binits{R.}}:
Gated-attention readers for text comprehension.
arXiv preprint arXiv:1606.01549
(2016)
\end{botherref}
\endbibitem

%%% 25
\bibitem[\protect\citeauthoryear{Olabiyi and Mueller}{2019}]{olabiyi2019dlgnet}
\begin{botherref}
\oauthor{\bsnm{Olabiyi}, \binits{O.}},
\oauthor{\bsnm{Mueller}, \binits{E.T.}}:
Dlgnet: a transformer-based model for dialogue response generation.
arXiv preprint arXiv:1908.01841
(2019)
\end{botherref}
\endbibitem

%%% 26
\bibitem[\protect\citeauthoryear{Lewis et~al.}{2020}]{lewis2020retrieval}
\begin{barticle}
\bauthor{\bsnm{Lewis}, \binits{P.}},
\bauthor{\bsnm{Perez}, \binits{E.}},
\bauthor{\bsnm{Piktus}, \binits{A.}},
\bauthor{\bsnm{Petroni}, \binits{F.}},
\bauthor{\bsnm{Karpukhin}, \binits{V.}},
\bauthor{\bsnm{Goyal}, \binits{N.}},
\bauthor{\bsnm{K{\"u}ttler}, \binits{H.}},
\bauthor{\bsnm{Lewis}, \binits{M.}},
\bauthor{\bsnm{Yih}, \binits{W.-t.}},
\bauthor{\bsnm{Rockt{\"a}schel}, \binits{T.}}, \betal:
\batitle{Retrieval-augmented generation for knowledge-intensive nlp tasks}.
\bjtitle{Advances in neural information processing systems}
\bvolume{33},
\bfpage{9459}--\blpage{9474}
(\byear{2020})
\end{barticle}
\endbibitem

%%% 27
\bibitem[\protect\citeauthoryear{Yao et~al.}{2023}]{yao2023llm}
\begin{botherref}
\oauthor{\bsnm{Yao}, \binits{J.-Y.}},
\oauthor{\bsnm{Ning}, \binits{K.-P.}},
\oauthor{\bsnm{Liu}, \binits{Z.-H.}},
\oauthor{\bsnm{Ning}, \binits{M.-N.}},
\oauthor{\bsnm{Liu}, \binits{Y.-Y.}},
\oauthor{\bsnm{Yuan}, \binits{L.}}:
Llm lies: Hallucinations are not bugs, but features as adversarial examples.
arXiv preprint arXiv:2310.01469
(2023)
\end{botherref}
\endbibitem

%%% 28
\bibitem[\protect\citeauthoryear{Ji et~al.}{2024}]{ji2024llm}
\begin{botherref}
\oauthor{\bsnm{Ji}, \binits{Z.}},
\oauthor{\bsnm{Chen}, \binits{D.}},
\oauthor{\bsnm{Ishii}, \binits{E.}},
\oauthor{\bsnm{Cahyawijaya}, \binits{S.}},
\oauthor{\bsnm{Bang}, \binits{Y.}},
\oauthor{\bsnm{Wilie}, \binits{B.}},
\oauthor{\bsnm{Fung}, \binits{P.}}:
Llm internal states reveal hallucination risk faced with a query.
arXiv preprint arXiv:2407.03282
(2024)
\end{botherref}
\endbibitem

%%% 29
\bibitem[\protect\citeauthoryear{Martino et~al.}{2023}]{martino2023knowledge}
\begin{bchapter}
\bauthor{\bsnm{Martino}, \binits{A.}},
\bauthor{\bsnm{Iannelli}, \binits{M.}},
\bauthor{\bsnm{Truong}, \binits{C.}}:
\bctitle{Knowledge injection to counter large language model (llm) hallucination}.
In: \bbtitle{European Semantic Web Conference},
pp. \bfpage{182}--\blpage{185}
(\byear{2023}).
\bcomment{Springer}
\end{bchapter}
\endbibitem

%%% 30
\bibitem[\protect\citeauthoryear{Li et~al.}{2024}]{li2024enhancing}
\begin{botherref}
\oauthor{\bsnm{Li}, \binits{J.}},
\oauthor{\bsnm{Yuan}, \binits{Y.}},
\oauthor{\bsnm{Zhang}, \binits{Z.}}:
Enhancing llm factual accuracy with rag to counter hallucinations: A case study on domain-specific queries in private knowledge-bases.
arXiv preprint arXiv:2403.10446
(2024)
\end{botherref}
\endbibitem

%%% 31
\bibitem[\protect\citeauthoryear{Izacard et~al.}{2023}]{izacard2023atlas}
\begin{barticle}
\bauthor{\bsnm{Izacard}, \binits{G.}},
\bauthor{\bsnm{Lewis}, \binits{P.}},
\bauthor{\bsnm{Lomeli}, \binits{M.}},
\bauthor{\bsnm{Hosseini}, \binits{L.}},
\bauthor{\bsnm{Petroni}, \binits{F.}},
\bauthor{\bsnm{Schick}, \binits{T.}},
\bauthor{\bsnm{Dwivedi-Yu}, \binits{J.}},
\bauthor{\bsnm{Joulin}, \binits{A.}},
\bauthor{\bsnm{Riedel}, \binits{S.}},
\bauthor{\bsnm{Grave}, \binits{E.}}:
\batitle{Atlas: Few-shot learning with retrieval augmented language models}.
\bjtitle{Journal of Machine Learning Research}
\bvolume{24}(\bissue{251}),
\bfpage{1}--\blpage{43}
(\byear{2023})
\end{barticle}
\endbibitem

%%% 32
\bibitem[\protect\citeauthoryear{Shi et~al.}{2023}]{shi2023replug}
\begin{botherref}
\oauthor{\bsnm{Shi}, \binits{W.}},
\oauthor{\bsnm{Min}, \binits{S.}},
\oauthor{\bsnm{Yasunaga}, \binits{M.}},
\oauthor{\bsnm{Seo}, \binits{M.}},
\oauthor{\bsnm{James}, \binits{R.}},
\oauthor{\bsnm{Lewis}, \binits{M.}},
\oauthor{\bsnm{Zettlemoyer}, \binits{L.}},
\oauthor{\bsnm{Yih}, \binits{W.-t.}}:
Replug: Retrieval-augmented black-box language models.
arXiv preprint arXiv:2301.12652
(2023)
\end{botherref}
\endbibitem

%%% 33
\bibitem[\protect\citeauthoryear{Siriwardhana et~al.}{2023}]{siriwardhana2023improving}
\begin{barticle}
\bauthor{\bsnm{Siriwardhana}, \binits{S.}},
\bauthor{\bsnm{Weerasekera}, \binits{R.}},
\bauthor{\bsnm{Wen}, \binits{E.}},
\bauthor{\bsnm{Kaluarachchi}, \binits{T.}},
\bauthor{\bsnm{Rana}, \binits{R.}},
\bauthor{\bsnm{Nanayakkara}, \binits{S.}}:
\batitle{Improving the domain adaptation of retrieval augmented generation (rag) models for open domain question answering}.
\bjtitle{Transactions of the Association for Computational Linguistics}
\bvolume{11},
\bfpage{1}--\blpage{17}
(\byear{2023})
\end{barticle}
\endbibitem

%%% 34
\bibitem[\protect\citeauthoryear{Herzig et~al.}{2020}]{herzig2020tapas}
\begin{botherref}
\oauthor{\bsnm{Herzig}, \binits{J.}},
\oauthor{\bsnm{Nowak}, \binits{P.K.}},
\oauthor{\bsnm{M{\"u}ller}, \binits{T.}},
\oauthor{\bsnm{Piccinno}, \binits{F.}},
\oauthor{\bsnm{Eisenschlos}, \binits{J.M.}}:
Tapas: Weakly supervised table parsing via pre-training.
arXiv preprint arXiv:2004.02349
(2020)
\end{botherref}
\endbibitem

%%% 35
\bibitem[\protect\citeauthoryear{Xu et~al.}{2020}]{xu2020layoutlm}
\begin{bchapter}
\bauthor{\bsnm{Xu}, \binits{Y.}},
\bauthor{\bsnm{Li}, \binits{M.}},
\bauthor{\bsnm{Cui}, \binits{L.}},
\bauthor{\bsnm{Huang}, \binits{S.}},
\bauthor{\bsnm{Wei}, \binits{F.}},
\bauthor{\bsnm{Zhou}, \binits{M.}}:
\bctitle{Layoutlm: Pre-training of text and layout for document image understanding}.
In: \bbtitle{Proceedings of the 26th ACM SIGKDD International Conference on Knowledge Discovery \& Data Mining},
pp. \bfpage{1192}--\blpage{1200}
(\byear{2020})
\end{bchapter}
\endbibitem

%%% 36
\bibitem[\protect\citeauthoryear{Song et~al.}{2022}]{song2022graph}
\begin{barticle}
\bauthor{\bsnm{Song}, \binits{Z.}},
\bauthor{\bsnm{Yang}, \binits{X.}},
\bauthor{\bsnm{Xu}, \binits{Z.}},
\bauthor{\bsnm{King}, \binits{I.}}:
\batitle{Graph-based semi-supervised learning: A comprehensive review}.
\bjtitle{IEEE Transactions on Neural Networks and Learning Systems}
\bvolume{34}(\bissue{11}),
\bfpage{8174}--\blpage{8194}
(\byear{2022})
\end{barticle}
\endbibitem

%%% 37
\bibitem[\protect\citeauthoryear{Howard and Ruder}{2018}]{howard2018universal}
\begin{botherref}
\oauthor{\bsnm{Howard}, \binits{J.}},
\oauthor{\bsnm{Ruder}, \binits{S.}}:
Universal language model fine-tuning for text classification.
arXiv preprint arXiv:1801.06146
(2018)
\end{botherref}
\endbibitem

%%% 38
\bibitem[\protect\citeauthoryear{Hu et~al.}{2022}]{hu2022lora}
\begin{barticle}
\bauthor{\bsnm{Hu}, \binits{E.J.}},
\bauthor{\bsnm{Shen}, \binits{Y.}},
\bauthor{\bsnm{Wallis}, \binits{P.}},
\bauthor{\bsnm{Allen-Zhu}, \binits{Z.}},
\bauthor{\bsnm{Li}, \binits{Y.}},
\bauthor{\bsnm{Wang}, \binits{S.}},
\bauthor{\bsnm{Wang}, \binits{L.}},
\bauthor{\bsnm{Chen}, \binits{W.}}, \betal:
\batitle{Lora: Low-rank adaptation of large language models.}
\bjtitle{ICLR}
\bvolume{1}(\bissue{2}),
\bfpage{3}
(\byear{2022})
\end{barticle}
\endbibitem

%%% 39
\bibitem[\protect\citeauthoryear{Han et~al.}{2024}]{han2024parameter}
\begin{botherref}
\oauthor{\bsnm{Han}, \binits{Z.}},
\oauthor{\bsnm{Gao}, \binits{C.}},
\oauthor{\bsnm{Liu}, \binits{J.}},
\oauthor{\bsnm{Zhang}, \binits{J.}},
\oauthor{\bsnm{Zhang}, \binits{S.Q.}}:
Parameter-efficient fine-tuning for large models: A comprehensive survey.
arXiv preprint arXiv:2403.14608
(2024)
\end{botherref}
\endbibitem

%%% 40
\bibitem[\protect\citeauthoryear{Houlsby et~al.}{2019}]{houlsby2019parameter}
\begin{bchapter}
\bauthor{\bsnm{Houlsby}, \binits{N.}},
\bauthor{\bsnm{Giurgiu}, \binits{A.}},
\bauthor{\bsnm{Jastrzebski}, \binits{S.}},
\bauthor{\bsnm{Morrone}, \binits{B.}},
\bauthor{\bsnm{De~Laroussilhe}, \binits{Q.}},
\bauthor{\bsnm{Gesmundo}, \binits{A.}},
\bauthor{\bsnm{Attariyan}, \binits{M.}},
\bauthor{\bsnm{Gelly}, \binits{S.}}:
\bctitle{Parameter-efficient transfer learning for nlp}.
In: \bbtitle{International Conference on Machine Learning},
pp. \bfpage{2790}--\blpage{2799}
(\byear{2019}).
\bcomment{PMLR}
\end{bchapter}
\endbibitem

%%% 41
\bibitem[\protect\citeauthoryear{Zhang et~al.}{2024}]{zhang2024raft}
\begin{bchapter}
\bauthor{\bsnm{Zhang}, \binits{T.}},
\bauthor{\bsnm{Patil}, \binits{S.G.}},
\bauthor{\bsnm{Jain}, \binits{N.}},
\bauthor{\bsnm{Shen}, \binits{S.}},
\bauthor{\bsnm{Zaharia}, \binits{M.}},
\bauthor{\bsnm{Stoica}, \binits{I.}},
\bauthor{\bsnm{Gonzalez}, \binits{J.E.}}:
\bctitle{Raft: Adapting language model to domain specific rag}.
In: \bbtitle{First Conference on Language Modeling}
(\byear{2024})
\end{bchapter}
\endbibitem

%%% 42
\bibitem[\protect\citeauthoryear{Es et~al.}{2024}]{es2024ragas}
\begin{bchapter}
\bauthor{\bsnm{Es}, \binits{S.}},
\bauthor{\bsnm{James}, \binits{J.}},
\bauthor{\bsnm{Anke}, \binits{L.E.}},
\bauthor{\bsnm{Schockaert}, \binits{S.}}:
\bctitle{Ragas: Automated evaluation of retrieval augmented generation}.
In: \bbtitle{Proceedings of the 18th Conference of the European Chapter of the Association for Computational Linguistics: System Demonstrations},
pp. \bfpage{150}--\blpage{158}
(\byear{2024})
\end{bchapter}
\endbibitem

\end{thebibliography}

\end{document}